\documentclass[]{jair}
\usepackage{algorithmicx}
\usepackage{algorithm}
\usepackage{algpseudocode}

\usepackage{array}
\usepackage{textcomp}
\usepackage{stfloats}
\usepackage{url}
\usepackage{verbatim}
\usepackage{graphicx}
\usepackage{float} 
\usepackage{subfigure}
\usepackage{bm}
\usepackage{todonotes}
\usepackage{booktabs}
\usepackage{siunitx}
\usepackage{multirow}
\usepackage{balance}
\usepackage{bm}
\usepackage{xcolor}
\usepackage{colortbl}
\usepackage[switch]{lineno}

\setcopyright{cc}
\copyrightyear{2026}
\acmDOI{10.1613/jair.1.20498}

\JAIRAE{Roberta Calegari}
\JAIRTrack{Fairness and Bias in AI}
\acmVolume{85}
\acmArticle{20}
\acmMonth{02}
\acmYear{2026}

\RequirePackage[
  datamodel=acmdatamodel,
  style=acmauthoryear,
  backend=biber,
  giveninits=true,
  uniquename=init
  ]{biblatex}

\begin{document}

\title[Procedural Fairness in Machine Learning]{Procedural Fairness in Machine Learning}

\author{Ziming Wang}
\orcid{0000-0002-3118-8742}
\email{wangzm2021@mail.sustech.edu.cn}
\affiliation{
  \institution{Guangdong Provincial Key Laboratory of Brain-inspired Intelligent Computation, Department of Computer Science and Engineering, Southern University of Science and Technology}
  \city{Shenzhen}
  \country{China}
  \postcode{518055}
}

\author{Changwu Huang}
\authornote{Corresponding authors}
\orcid{0000-0003-3685-2822}
\email{changwuhuang@bnbu.edu.cn}
\affiliation{
  \institution{School of AI and Liberal Arts, Beijing Normal-Hong Kong Baptist University}
  \city{Zhuhai}
  \country{China}
  \postcode{519087}
}

\author{Ke Tang}
\authornotemark[1]
\orcid{0000-0002-6236-2002}
\email{tangk3@sustech.edu.cn}
\affiliation{
  \institution{Guangdong Provincial Key Laboratory of Brain-inspired Intelligent Computation, Department of Computer Science and Engineering, Southern University of Science and Technology}
  \city{Shenzhen}
  \country{China}
  \postcode{518055}
}

\author{Xin Yao}
\orcid{0000-0001-8837-4442}
\email{xinyao@ln.edu.hk}
\affiliation{
  \institution{School of Data Science, Lingnan University}
  \country{Hong Kong, China}
}

\renewcommand{\shortauthors}{Wang, Huang, Tang \& Yao}

\begin{abstract}
Fairness in machine learning (ML) has garnered significant attention. However, current research has mainly concentrated on the distributive fairness of ML models, with limited focus on another dimension of fairness, i.e., procedural fairness. In this paper, we first define the procedural fairness of ML models by drawing from the established understanding of procedural fairness in philosophy and psychology fields, and then give formal definitions of individual and group procedural fairness. Based on the proposed definition, we further propose a novel metric to evaluate the group procedural fairness of ML models, called \textit{GPF\textsubscript{FAE}}, which utilizes a widely used explainable artificial intelligence technique, namely feature attribution explanation (FAE), to capture the decision process of ML models. We validate the effectiveness of \textit{GPF\textsubscript{FAE}} on a synthetic dataset and eight real-world datasets. Our experimental studies have revealed the relationship between procedural and distributive fairness of ML models. After validating the proposed metric for assessing the procedural fairness of ML models, we then propose a method for identifying the features that lead to the procedural unfairness of the model and propose two methods to improve procedural fairness based on the identified unfair features. Our experimental results demonstrate that we can accurately identify the features that lead to procedural unfairness in the ML model, and both of our proposed methods can significantly improve procedural fairness while also improving distributive fairness, with a slight sacrifice on the model performance.
\end{abstract}

\received{22 September 2025}
\received[accepted]{10 January 2026}

\maketitle

\section{Introduction}
\label{Introduction}

As artificial intelligence (AI) is increasingly used in critical domains such as finance~\cite{chen2016financial}, hiring~\cite{li2021hiring}, and criminal justice~\cite{dressel2018accuracy} to make consequential decisions affecting individuals, concerns about discrimination and fairness inevitably arise and have become the forefront of deliberations within the realm of AI ethics~\cite{9844014,li2021trust}. Generally, in the context of decision-making, fairness refers to the “\textit{absence of any prejudice or favoritism toward an individual or a group based on their inherent or acquired characteristics}''~\cite{mehrabi2021survey}. This is a commonly accepted conception of fairness in AI, especially in fairness-aware machine learning (ML). The two well-known dimensions of fairness are distributive fairness (a.k.a outcome fairness) and procedural fairness (a.k.a process fairness)~\cite{greenberg1987taxonomy,morse2021ends}. Among them, distributive fairness focuses on the fairness of outcomes resulting from a decision, while procedural fairness emphasizes fairness in the decision-making process~\cite{morse2021ends,grgic2018beyond}. There has long been considerable discussion in the humanities community about the relationship and distinction between the two~\cite{ambrose2013procedural,folger1987distributive}. Although distributive and procedural fairness mutually influence justice assessments, procedural fairness is considered to be the more robust assessment criterion~\cite{morse2021ends,thibaut1975procedural,lind2001heuristic}. Further, people are more willing to support an unfair outcome when they feel the process is fair~\cite{morse2021ends,van1998we}.

In the last decades, numerous studies have explored fairness in ML, and various quantitative fairness metrics~\cite{fabris2023measuring} and unfairness mitigation approaches have been proposed~\cite{mehrabi2021survey,pessach2022review,blandin2023generalizing}. However, these efforts are primarily undertaken from the perspective of distributive fairness and do not adequately address procedural fairness in ML~\cite{morse2021ends,grgic2018beyond}. Until now, relatively little attention has been paid to procedural fairness~\cite{mehrabi2021survey,morse2021ends,grgic2018beyond,zhao2022fairness}, and even a better definition of procedural fairness in ML models is lacking~\cite{balkir2022challenges}.

Notably, in the philosophy and social sciences literature, procedural fairness has been well studied, emphasizing whether the procedures, rules, and processes used in decision-making are fair, transparent, and equitable~\cite{thibaut1975procedural,leventhal1980should}. Inspired by this perspective, we extend and adapt these concepts into the context of ML. This adaptation ensures consistency with established procedural fairness concepts while focusing on the decision-making processes of ML models. Moreover, while major AI governance and standardization initiatives (e.g., the EU AI Act~\cite{EU_AI_Act_2024}, OECD AI Principles~\cite{oecd2024aiprinciples}) emphasize fairness as a core principle, they generally do not explicitly distinguish between distributive and procedural fairness. In practice, most existing technical implementations and fairness metrics have focused predominantly on distributive outcomes. Our work addresses this imbalance by introducing a formal definition and a quantifiable metric for procedural fairness, offering a decision-process-oriented perspective that complements existing technical efforts and can be aligned with broader fairness objectives outlined in these governance frameworks.

To the best of our knowledge, the only work that has both defined and evaluated the procedural fairness of the ML model is Grgi{\'c}-Hla{\v{c}}a \textit{et al.}~\citeyear{grgic2018beyond}, in which the authors defined the procedural fairness of a model in terms of the inherent fairness of the features used in training the model. However, no equivalence can be drawn between the fairness of the features actually used and the procedural fairness of the ML model. For example, an ML model obtained by training with unfair features (e.g., sensitive features) does not imply that its decision-making process is unfair, as verified by our experimental results on the \textit{COMPAS} dataset in Section \ref{sec:definition_and_metric} as a counterexample. The key problem with this is that their definition solely perceives the procedural fairness of the ML model based on the input features, without considering whether the decision-making process or logic behind the model's predictions is fair or not. According to the concept of procedural fairness~\cite{morse2021ends}, the decision-making process is indeed the core element and cornerstone in perceiving procedural fairness. From this perspective, it is evident that this definition does not accurately capture the key essence of procedural fairness. Furthermore, as they themselves stated~\cite{grgic2018beyond}, they measured procedural fairness by “\textit{relying on humans' moral judgments or instincts of humans about the fairness of using input features in a decision-making context}.'' While incorporating human perspectives is essential in defining fairness, such evaluation methods come with significant practical challenges: they are costly to implement, difficult to scale to new tasks or datasets, and may yield inconsistent results across different populations or settings. In summary, the urgency of research around procedural fairness (including its conceptual definition, evaluation methods, and related aspects) is evident. However, within the overall field of AI fairness, the study of procedural fairness is currently in its infancy, highlighting the need for more and enhanced research efforts.

This motivates us to focus our efforts on procedural fairness in ML. In this paper, we present a systematic approach to defining, evaluating, and improving procedural fairness in ML models. Building on prior understanding from the social sciences, we first provide a clear definition of procedural fairness in the context of ML and formally define individual and group procedural fairness. Then, based on the proposed definition, a novel quantitative metric to assess group procedural fairness is proposed based on feature attribution explanation (FAE)~\cite{guidotti2018survey}, which is a popular and well-studied explainable AI (XAI) method that explains decisions of ML models by calculating the attribution of each input feature (i.e., the importance score of each input feature)~\cite{wang2024roadmap,bhatt2020evaluating,wang2023feature,wang2024multi}. Our proposed procedural fairness metric can provide insight into the decision process of the model through the FAE method and evaluate the group procedural fairness of the model by assessing the overall distribution difference of FAE explanations between similar data points of two groups. To evaluate the effectiveness of our proposed definitions and methods, nine different datasets are used in our experimental studies. The relationship between procedural fairness and distributive fairness of the ML model is further analyzed. Finally, in the case where the model exhibits procedural unfairness, we have developed an approach that traces the source of procedural unfairness by identifying features that are significantly different in the FAE explanations of the two groups, and propose two mitigation methods to improve procedural fairness based on the identified unfair features. Our experimental results on the nine datasets show that our methods are indeed able to find the features that lead to procedural unfairness accurately, and that both mitigation methods significantly improve procedural fairness with a slight impact on model performance, while also improving distributive fairness at the same time.

The main contributions of this paper include the following:
\begin{enumerate}
    \item[(1)] We provide a more precise and more comprehensive definition of procedural fairness for ML models by drawing on existing research in the humanities, and further define individual and group procedural fairness formally.
    \item[(2)] We propose a quantitative metric based on FAE to assess the group procedural fairness based on the proposed definition, and validate the metric through experimental studies. Furthermore, we analyze the relationship between procedural fairness and distributive fairness of the ML model.
    \item[(3)] We propose a detection method to identify the sources of unfairness. For a procedural unfairness ML model, the features that lead to procedural unfairness are found by detecting features that are significantly different in the FAE explanations. Then, based on the identified unfair features, we propose two approaches to mitigate procedural unfairness. Experimental results on nine datasets show that our proposed detection method can accurately identify the features contributing to procedural unfairness, and both proposed enhancement methods can effectively improve procedural fairness without significantly compromising the model's performance, while simultaneously improving distributive fairness.
\end{enumerate}

The rest of this paper is organized as follows. Section \ref{sec:related_work} presents related work. Section \ref{sec:definition_and_metric} defines the procedural fairness of the ML model, and then proposes a metric to assess group procedural fairness and evaluates its effectiveness. Section \ref{sec:identify_improve} proposes a method to identify the features that lead to procedural unfairness in the model, and two methods to improve procedural fairness are proposed and experimentally studied. Section \ref{sec:conclusion} concludes the paper and gives future research directions.

\section{Related Work}
\label{sec:related_work}

In this section, we briefly review the background and related work on fairness in ML. We first introduce the definitions and measurements of distributive fairness, which have been extensively studied in the literature. Then, we review existing work on the procedural fairness of ML models, which has received less attention. Finally, we present related work that has applied the FAE approach to fairness.

\subsection{Related Work on Distributive Fairness in ML}
\label{sec:distributive_fairness_ml}
In recent years, fairness in ML has attracted increasing attention~\cite{pessach2022review}, and researchers have proposed various definitions and metrics to measure the distributive fairness of ML models, which can be categorized into individual and group distributive fairness~\cite{mehrabi2021survey}. 

\textbf{Individual distributive fairness} refers to ML models that make similar predictions for similar individuals~\cite{mehrabi2021survey}, and its formal definition is described below.
\begin{itemize}
    \item \textbf{Individual Fairness (IF)}~\cite{dwork2012fairness,benussi2022individual}: this measure requires that similar individuals should be treated similarly~\cite{pessach2022review}. Similarity can be defined with respect to a particular task~\cite{dwork2012fairness}. \textit{IF} can be described as:
\begin{equation}
\label{eq:IF}
IF = |\hat{y} ^{(i)}-\hat{y} ^{(j)}|; \;\; s.t. \;\; d(\textbf{\textit{x}}^{(i)},\textbf{\textit{x}}^{(j)})\leq \varepsilon,
\end{equation}
where $\textbf{\textit{x}}^{(\cdot)}$ denotes an individual, $\hat{y}^{(\cdot)}$ refers to the corresponding label predicted by the ML model for the individual, $d(\cdot,\cdot)$ is a distance metric between two individuals, and $\varepsilon$ is a threshold to control the similarity between individuals.
\end{itemize}
\textbf{Group distributive fairness} refers to ML models that treat different groups equally~\cite{mehrabi2021survey}, and there are three common metrics, namely Demographic Parity (DP)~\cite{dwork2012fairness,calders2010three}, Equal Opportunity (EO)~\cite{hardt2016equality}, and Equalized Odds (EOD)~\cite{hardt2016equality}.

\begin{itemize}
    \item \textbf{Demographic Parity (DP)}~\cite{dwork2012fairness,calders2010three}: this measure requires that the difference between the positive prediction rates of different sensitive groups should be as small as possible. It is commonly referred to as statistical parity. \textit{DP} can be described as:
\begin{equation}
\label{eq:DP}
DP = \left | P\left(\hat{y}=1 | s=s_{1}\right) - P\left(\hat{y}=1 | s=s_{2}\right) \right |,
\end{equation}
where $\hat{y}$ is the predicted labels of the classifier, $s$ represents the sensitive attribute (e.g., sex, race), and $s_1$ and $s_2$ denote two different groups associated with the sensitive attribute.
\end{itemize}
\begin{itemize}
    \item \textbf{Equal Opportunity (EO)}~\cite{hardt2016equality}: this measure requires that the difference between the true-positive rates of different sensitive groups should be as small as possible. \textit{EO} can be described as:
\begin{equation}
\label{eq:EO}
EO= \left | P\left(\hat{y}=1 | s=s_{1}, y=1\right) - P\left(\hat{y}=1 | s=s_{2}, y=1\right) \right |,
\end{equation}
where $y$ is the true label, and the definitions of the other symbols are the same as for the \textit{DP} metric.
\end{itemize}

\begin{itemize}
    \item \textbf{Equalized Odds (EOD)}~\cite{hardt2016equality}: this measure requires that both the differences between the false-positive rates and the true-positive rates of different sensitive groups should be as small as possible. \textit{EOD} can be described as:
\end{itemize}
{\small
\begin{equation}
\label{eq:EOD}
\begin{split}
EOD = \frac{1}{2}&\times\left (| P\left(\hat{y}=1 | s=s_{1}, y=0\right) - P\left(\hat{y}=1 | s=s_{2}, y=0\right) \right |  \\ 
        & + \left | P\left(\hat{y}=1 | s=s_{1}, y=1\right) - P\left(\hat{y}=1 | s=s_{2}, y=1\right) \right |).
\end{split}
\end{equation}}

To improve the distributive fairness of ML models, researchers have proposed various pre-processing~\cite{zemel2013learning,kamiran2012data}, in-processing~\cite{9902997,zhang2021fairer,zhang2018mitigating}, and post-processing~\cite{hardt2016equality} mechanisms. Although these methods are valuable in mitigating bias in prediction outcomes, existing fairness metrics and mitigation methods mainly focus on distributive fairness while neglecting procedural fairness~\cite{zhao2022fairness}.

\subsection{Related Work on Procedural Fairness in ML}
\label{sec:rw_pfml}

Currently, there are only some survey papers that discuss procedural fairness in ML~\cite{mehrabi2021survey,morse2021ends,pessach2022review,guidotti2018survey,tang2023and,green2018myth}, and many of them point out that there is little attention paid to procedural fairness~\cite{mehrabi2021survey,morse2021ends,grgic2018beyond,zhao2022fairness}. A notable exception and a very important work is Grgi{\'c}-Hla{\v{c}}a \textit{et al.}~\citeyear{grgic2018beyond}, which defined the procedural fairness of the ML model by the fairness of the features it uses, i.e., “\textit{We consider a classifier $C$ trained using a subset of features $F$ from a set $\bar{F} $ of all possible features. We define the process fairness of $C$ to be the fraction of all users who consider the use of every feature in $F$ to be fair}''. However, as mentioned before, this definition assesses the procedural fairness of a model based only on the inherent fairness of its input features, without considering the decision-making process or logic inside the model. Additionally, their evaluation method relies on human assessments, which are costly to conduct and difficult to scale to new tasks or datasets. Moreover, their definition equates the procedural fairness of ML models with the fairness of the features used. However, a model that uses unfair features (e.g., sensitive features) does not imply that its decision process is unfair. This is illustrated by our experiments on the \textit{COMPAS} dataset in Section \ref{sec:definition_and_metric} as a counterexample. As pointed out by Balkir \textit{et al.}~\citeyear{balkir2022challenges}, procedural fairness lacks a better definition as well as precise quantitative metrics similar to those that have been developed for distributive fairness.

In this study, we establish the definition of procedural fairness of the ML model by considering the consistency of decision logic across individuals or groups. That is, our proposed procedural fairness definition captures the fairness of the model's decision process, unlike distributive fairness, which considers the equality of the model's prediction (output) between different individuals or groups. 

\subsection{Related Work on Applying FAE to Fairness}

XAI technologies attempt to provide understandable explanations from different perspectives for humans to gain insight into the decisions of AI systems~\cite{wang2024roadmap,arrieta2020explainable,guidotti2018survey}. One kind of such techniques, known as feature attribution explanation (FAE) method, explains the predictions of an ML model by computing the attribution of each input feature (i.e., the importance of each input feature)~\cite{bhatt2020evaluating,wang2024multi}, which helps to identify the most significant features influencing the model's predictions, thus providing valuable insights into the decision-making procedure and rationales behind the ML model's decision~\cite{zhao2022fairness}. 

In this paper, we primarily employ SHAP~\cite{lundberg2017unified}, a popular perturbation-based FAE method, to gain insight into the decision process of the ML models. Its advantage lies in its model-agnostic nature, as it does not require internal model information and can be applied to explain any ML model. To verify the robustness and generality of our approach, we further incorporate two additional FAE techniques: gradient*input (GI)~\cite{shrikumar2016not} and integrated gradients (IG)~\cite{sundararajan2017axiomatic}, which are representative gradient-based explanation methods. Their advantage lies in their extremely high computational efficiency. By comparing results across multiple FAE methods, we ensure that our fairness evaluations are not overly reliant on a specific explanation technique and are consistent across diverse explanation perspectives.

Although FAE is the most widely used XAI technique~\cite{arrieta2020explainable,bhatt2020explainable}, and much of the literature points out that XAI plays an important role in achieving ML fairness~\cite{balkir2022challenges,abdollahi2018transparency}, only a few works apply the FAE methods to fairness~\cite{zhao2022fairness,dai2022fairness}. Among them, Begley \textit{et al.}~\citeyear{begley2020explainability} and Pan \textit{et al.}~\citeyear{pan2021explaining} used FAE methods to attribute distributive fairness metrics to capture the contribution of each feature to an unfair decision result. Dimanov \textit{et al.}~\citeyear{dimanoyouv2020you} modified the ML model to reduce the importance score of the sensitive feature obtained by FAE and found that the model was still unfair, thus pointing out that direct observation of the importance score of the sensitive feature obtained by FAE does not reliably reveal the fairness of the model. The most relevant research to us is the work of Zhao \textit{et al.}~\citeyear{zhao2022fairness}, which utilized FAE to capture insights into the model's decision process and to identify bias in a procedural-oriented manner. However, their focus is on the differences between the quality of explanations obtained by different groups, i.e., the fairness of the explanations rather than the procedural fairness of the model's predictions. 

In short, although many researchers have pointed out that XAI could improve the fairness of ML models~\cite{balkir2022challenges,abdollahi2018transparency}, there is no existing work that utilizes FAE (or other XAI techniques) to assess the fairness of the ML model. In this paper, we evaluate and improve the procedural fairness of the ML model based on the FAE method.

\section{New Definition and Measurement of Procedural Fairness}
\label{sec:definition_and_metric}

In this section, we first introduce the relevant notations used in this paper. Then, we provide a definition of procedural fairness for the ML model and propose an FAE-based measurement to assess the group procedural fairness. Subsequently, the effectiveness of the proposed procedural fairness metric is validated through extensive experiments. Furthermore, the relationship between procedural fairness and distributive fairness is analyzed. Lastly, we point out some limitations of our proposed procedural fairness measurement and offer promising solutions.

\subsection{Notation}

We consider the binary classification problem that aims to learn a mapping function between input feature vectors $\textit{\textbf{x}}\in\mathbb{R}^{d}$ and class labels ${y}\in \{0, 1\}$ based on a given dataset $\bm{\bm{\mathcal{D}}}=(\textit{\textbf{X}},\textit{\textbf{Y}})= \{ (\textit{\textbf{x}}^{(i)},y^{(i)})\}_{i=1}^{m}$, where $\textit{\textbf{x}}^{(i)}=[ x_{1}^{(i)},\cdots, x_{d}^{(i)}] \in  \mathbb{R}^{d} $ are feature vectors and $y^{(i)} \in \{0, 1\}$ are their corresponding labels. This task is often achieved by finding a model or classifier $f: \textit{\textbf{x}} \mapsto {y}$ based on the dataset $\bm{\bm{\mathcal{D}}}$ so that given a feature vector $\textit{\textbf{x}}$ with unknown label $y$, the classifier can predict its label $\hat{y}=f\left (\textit{\textbf{x}}\right )$. Also, each data point $\textit{\textbf{x}}$ has an associated sensitive attribute $s$ (e.g., sex, race) that indicates the group membership of an individual. Actually, there can be multiple sensitive attributes, but in this paper, we consider, without loss of generality, a single sensitive attribute case (e.g., the gender of each user $s=\{male, female\}$) and use $s_1$ and $s_2$ to denote two different groups associated with the sensitive attribute. That is, each data point $(\textit{\textbf{x}}^{(i)},y^{(i)})\in\bm{\bm{\mathcal{D}}}$ has an associated sensitive feature value $s^{(i)}\in\{s_1,s_2\}$. Correspondingly, the subsets or groups of dataset $\bm{\bm{\mathcal{D}}}$ with values $s=s_1$ and $s=s_2$ are denoted as $\bm{\mathcal{D}}_1=(\textit{\textbf{X}}_1,\textit{\textbf{Y}}_1)= \{(\textit{\textbf{x}}^{(i)},y^{(i)})\in\bm{\mathcal{D}}|s^{(i)}=s_1\}$ and $\bm{\mathcal{D}}_2=(\textit{\textbf{X}}_2,\textit{\textbf{Y}}_2)= \{(\textit{\textbf{x}}^{(i)},y^{(i)})\in\bm{\mathcal{D}}|s^{(i)}=s_2\}$, respectively. 

For a model $f: \textit{\textbf{x}} \mapsto {\hat{y}}$, the symbol $\mapsto$ denotes the decision-making process or mapping process of the ML model $f$. However, for the vast majority of ML models, the mapping process is opaque, making it difficult to measure and assess its procedural fairness. Therefore, to formally define procedural fairness, we seek to establish a representation, denoted as $\Phi$, that portrays the decision-making process of the ML model. Unlike the mapping process $\mapsto$, $\Phi$ should be understandable, quantifiable, and comparable, allowing us to quantify and assess the procedural fairness of the ML model. Additionally, we use $g$ to represent a local FAE function which takes a model $f$ and an explained data point $\textit{\textbf{x}}^{(i)}$ as inputs and returns explanations (i.e., feature importance scores) $\textit{\textbf{e}}^{(i)}=g(f,\,\textit{\textbf{x}}^{(i)})\in\mathbb{R}^{d}$, where $e^{(i)}_j$ (i.e., $g(f,\,\textit{\textbf{x}}^{(i)})_j$) is the importance score of the feature $x^{(i)}_j$ for the model's prediction $f(\textit{\textbf{x}}^{(i)})$. 

\subsection{Definition of Procedural Fairness}
\label{sec:define}

In this subsection, we establish a definition of procedural fairness in ML models by drawing on research on procedural fairness in philosophy~\cite{thibaut1975procedural,leventhal1980should} and fairness concepts in the field of ML~\cite{mehrabi2021survey}. Specifically, we focus on the decision logic of ML models and extend the notion of ML fairness to encompass procedural elements, and further formally define individual and group procedural fairness. 

\noindent\textbf{Definition 1} The procedural fairness of the ML model is defined as the internal decision process or logic of the model without any prejudice or preference for individuals or groups due to their inherent or acquired characteristics.

\noindent\textbf{Definition 2} The individual procedural fairness of the ML model is defined as that two similar data points $\textit{\textbf{x}}^{(i)}$ and $\textit{\textbf{x}}^{(j)}$ should have similar decision processes or logic:
\begin{equation}
\label{eq:IPF}
d_{\Phi}(\Phi(\textit{\textbf{x}}^{(i)}),\Phi(\textit{\textbf{x}}^{(j)}))\approx 0\;\;
\text{s.t.} \;\; d_{\textit{\textbf{x}}}(\textbf{\textit{x}}^{(i)},\textbf{\textit{x}}^{(j)})\approx 0,
\end{equation}
where $d_{\Phi}$ and $d_{\textit{\textbf{x}}}$ are used to measure the similarity of two decision processes and two data points, respectively.

\noindent\textbf{Definition 3} The group procedural fairness of the ML model is defined as that similar data points in two groups $\textit{\textbf{x}}^{(i)}\in\textit{\textbf{X}}_1$ and $\textit{\textbf{x}}^{(j)}\in\textit{\textbf{X}}_2$ should have similar decision processes or logic:

\begin{equation}
\label{eq:GPF}
d_{\Phi}(\Phi(\textit{\textbf{x}}^{(i)}),\Phi(\textit{\textbf{x}}^{(j)}))\approx 0\;\;
 \text{s.t.} \;\; \textit{\textbf{x}}^{(i)}\in\textit{\textbf{X}}_1, \; \textit{\textbf{x}}^{(j)}\in\textit{\textbf{X}}_2 \;
\text{and} \; d_{\textit{\textbf{x}}}(\textbf{\textit{x}}^{(i)},\textbf{\textit{x}}^{(j)})\approx 0.
\end{equation}

Consistent with individual distributive fairness, similarity $d_{\Phi}$ and $d_{\textit{\textbf{x}}}$ in Definitions 2 and 3 are often defined with respect to a particular task. The biggest challenge is to find a representation $\Phi$ that can be used to portray the ML models' decision logic or process. 

As discussed in Section \ref{sec:rw_pfml}, the difference between our definition of procedural fairness and Grgi{\'c}-Hla{\v{c}}a \textit{et al.}~\citeyear{grgic2018beyond} is that they defined procedural fairness in terms of the fairness of the features (based on human judgment) used by an ML model. Instead, we define the model's procedural fairness based on the model's decision logic or process. We further formally define procedural fairness for individuals and groups, capturing fairness at the individual and group levels, respectively.

\subsection{Measurement of Procedural Fairness}
\label{sec:metric}

Despite the definition of procedural fairness given, as mentioned before, the biggest challenge in assessing the procedural fairness of ML models is how to portray the ML models' decision logic or process $\Phi$. Fortunately, XAI techniques can provide insight into the decision logic within the ML model. In particular, Zhao \textit{et al.}~\citeyear{zhao2022fairness} used FAE techniques to capture the model's decision process for identifying biases and showed its effectiveness. Therefore, we also use FAE techniques to capture the model's decision process $\Phi$. Specifically, given a data point $\textit{\textbf{x}}^{(i)}$, its decision logic $\Phi(\textit{\textbf{x}}^{(i)})$ is portrayed as its FAE result $\textit{\textbf{e}}^{(i)}=g(f,\,\textit{\textbf{x}}^{(i)})$. Then we propose a quantitative FAE-based metric for assessing the group procedural fairness, called FAE-based Group Procedural Fairness (\textit{GPF\textsubscript{FAE}}). \textit{GPF\textsubscript{FAE}} metric can be regarded as an instantiation of Definition 3, which is defined as: 

\begin{equation}
    \begin{split}
        \textit{GPF\textsubscript{FAE}}=&d_{\Phi}(\textit{\textbf{E}}_1,\textit{\textbf{E}}_2);\\
        \textit{\textbf{E}}_1=&\{\textit{\textbf{e}}^{(i)}|\textit{\textbf{e}}^{(i)}\,=g(f, \textit{\textbf{x}}^{(i)}\,),\textit{\textbf{x}}^{(i)}\in \textit{\textbf{X}}^{'}_{1}\},\\
        \textit{\textbf{E}}_2=&\{\textit{\textbf{e}}^{(j)}|\textit{\textbf{e}}^{(j)}=g(f, \textit{\textbf{x}}^{(j)}),\textit{\textbf{x}}^{(j)}\in \textit{\textbf{X}}^{'}_{2}\}, \\
    \end{split}
\end{equation}

\noindent where $d_{\Phi}(\cdot, \cdot)$ is a measurement of the distance between two sets of FAE explanation results $\textit{\textbf{E}}_1$ and $\textit{\textbf{E}}_2$, $\textit{\textbf{X}}^{'}_{1}$ and $\textit{\textbf{X}}^{'}_{2}$ are sets of $n$ data points from $\textit{\textbf{X}}_{1}$ and $\textit{\textbf{X}}_{2}$, respectively, representing $n$ pairs of similar data points in the two groups, which are generated as shown in Algorithm~\ref{alg:select_dataset}.

\begin{algorithm}
    \caption{Selecting datasets for \textit{GPF\textsubscript{FAE}}.} 
    \label{alg:select_dataset}
    \begin{algorithmic}[1]
        \Require Two subsets $\textit{\textbf{X}}_{1}$ and $\textit{\textbf{X}}_{2}$ of the dataset $\textit{\textbf{X}}$, the total number of selected data points $n$, the data point similarity (or distance) metric $d_{\textit{\textbf{x}}}$.
        \Ensure The datasets $\textit{\textbf{X}}^{'}_{1}$ and $\textit{\textbf{X}}^{'}_{2}$ for evaluating \textit{GPF\textsubscript{FAE}}.
        \State $\textit{\textbf{X}}^{'}_{1}=\emptyset,\textit{\textbf{X}}^{'}_{2}=\emptyset$.
        \For {$k=1, \cdots,\lfloor \frac{n}{2}\rfloor$}
            \State Randomly select a data point $\textit{\textbf{x}}^{(i)}$ from $\textit{\textbf{X}}_{1}$.
            \State Find the $\textit{\textbf{x}}^{(j)} \in \textit{\textbf{X}}_{2}$ with the smallest $d_{\textit{\textbf{x}}}(\textit{\textbf{x}}^{(i)},\textit{\textbf{x}}^{(j)})$.
            \State $\textit{\textbf{X}}^{'}_{1}=\textit{\textbf{X}}^{'}_{1}\cup\textit{\textbf{x}}^{(i)}$, $\textit{\textbf{X}}^{'}_{2}=\textit{\textbf{X}}^{'}_{2}\cup\textit{\textbf{x}}^{(j)}$.
        \EndFor
        \For {$k=\lfloor \frac{n}{2}\rfloor+1, \cdots, n$}
            \State Randomly select a data point $\textit{\textbf{x}}^{(j)}$ from $\textit{\textbf{X}}_{2}$.
            \State Find the $\textit{\textbf{x}}^{(i)} \in \textit{\textbf{X}}_{1}$ with the smallest $d_{\textit{\textbf{x}}}(\textit{\textbf{x}}^{(j)},\textit{\textbf{x}}^{(i)})$.
            \State $\textit{\textbf{X}}^{'}_{1}=\textit{\textbf{X}}^{'}_{1}\cup\textit{\textbf{x}}^{(i)}$, $\textit{\textbf{X}}^{'}_{2}=\textit{\textbf{X}}^{'}_{2}\cup\textit{\textbf{x}}^{(j)}$.
        \EndFor
        \State \Return the datasets $\textit{\textbf{X}}^{'}_{1}$ and $\textit{\textbf{X}}^{'}_{2}$.
	\end{algorithmic} 
\end{algorithm}

The datasets $\textit{\textbf{X}}^{'}_{1}$ and $\textit{\textbf{X}}^{'}_{2}$ used to evaluate the procedural fairness metric \textit{GPF\textsubscript{FAE}} should be representative of each group involved in the dataset. To ensure this, $\textit{\textbf{X}}^{'}_{1}$ and $\textit{\textbf{X}}^{'}_{2}$ are created through two distinct phases. Given the two subsets of each group $\textit{\textbf{X}}_{1}$ and $\textit{\textbf{X}}_{2}$ from the dataset $\textit{\textbf{X}}$, the total number of selected data points $n$, and a similarity or distance metric $d_{\textit{\textbf{x}}}(\cdot, \cdot)$ between individual data points, datasets $\textit{\textbf{X}}^{'}_{1}$ and $\textit{\textbf{X}}^{'}_{2}$ with the size of $n$ are created as follows: in the first phase, for the first half of the dataset size $n$, a data point $\textit{\textbf{x}}^{(i)}$ is randomly chosen from $\textit{\textbf{X}}_{1}$  (Line 3). Subsequently, the data point $\textit{\textbf{x}}^{(j)} \in \textit{\textbf{X}}_{2}$ that has the smallest distance to $\textit{\textbf{x}}^{(i)}$ measured by $d_{\textit{\textbf{x}}}$ is selected  (Line 4). These two selected data points are then added to $\textit{\textbf{X}}^{'}_{1}$ and $\textit{\textbf{X}}^{'}_{2}$, respectively (Line 5). For the remaining half of the dataset size $n$, a data point $\textit{\textbf{x}}^{(j)}$ is randomly selected from $\textit{\textbf{X}}_{2}$ (Line 8). Following this, the data point $\textit{\textbf{x}}^{(i)} \in \textit{\textbf{X}}_{1}$ with the smallest distance $d_{\textit{\textbf{x}}}(\textit{\textbf{x}}^{(j)},\textit{\textbf{x}}^{(i)})$ is chosen (Line 9). The selected pair of data points is then collected in $\textit{\textbf{X}}^{'}_{1}$ and $\textit{\textbf{X}}^{'}_{2}$, respectively (Line 10). In this way, there are $n$ data points in each of $\textit{\textbf{X}}^{'}_{1}$ and $\textit{\textbf{X}}^{'}_{2}$ with the minimum distance between each pair of them.

In this paper, we use the maximum mean discrepancy (MMD) with an exponential kernel function as $d_{\Phi}$ to measure the distributional differences between two explanation sets $\textit{\textbf{E}}_1$ and $\textit{\textbf{E}}_2$, and the $p$-value obtained by performing a permutation test on the generated kernel matrix is used as the final evaluation result. If the $p$-value is larger than a threshold, we consider the model meets the procedural fairness requirement; otherwise, the model is deemed procedurally unfair. The Euclidean distance is used as the distance metric $d_{\textit{\textbf{x}}}$ for identifying the most similar data points. The ML models employed in this study are artificial neural networks (ANNs), although our method is applicable to any other ML models.

\begin{figure}[htbp]
    \centering
    \includegraphics[width=0.8\linewidth]{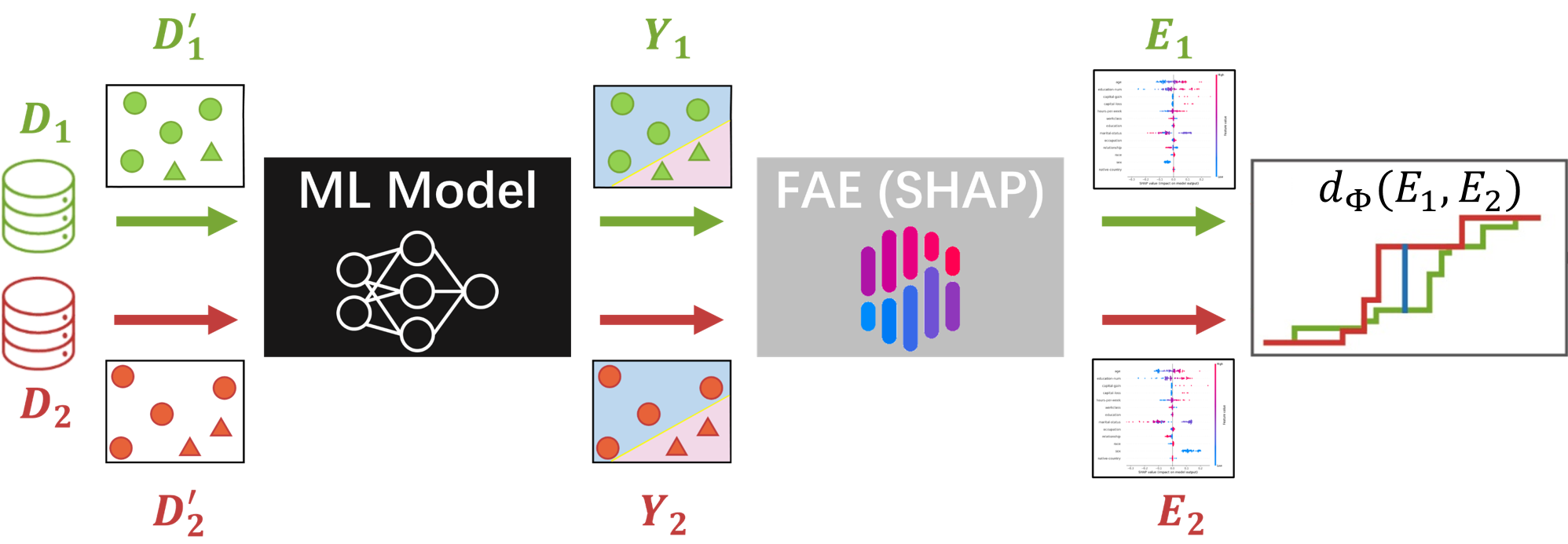}
    \caption{Pipeline of using \textit{GPF\textsubscript{FAE}} metric to evaluate group procedural fairness.}
    \label{fig:FAEEF_process}
\end{figure}

In short, the \textit{GPF\textsubscript{FAE}} quantifies the disparity in FAE explanation results between similar data points belonging to two different groups. The pipeline of using our proposed \textit{GPF\textsubscript{FAE}} to measure the group procedural fairness of the ML model is shown in Fig.~\ref{fig:FAEEF_process}. 

To simplify our discussions, we refer to an ML model that meets distributive fairness as a distributive-fair model, and a model that does not meet distributive fairness criteria as a distributive-unfair model. Similarly, an ML model that is in line with our defined group procedural fairness metric is denoted as a procedural-fair model, and a model that is not in line with procedural fairness criteria is called a procedural-unfair model. These terms will be used throughout the remainder of this paper.

\subsection{Experimental Evaluation of the Proposed Procedural Fairness Metric}
\label{sec:evaluate_metric}
In this subsection, we extensively validate and evaluate our proposed metric, \textit{GPF\textsubscript{FAE}}, through a comprehensive set of experimental studies.
We begin by presenting the relevant experimental setup, providing details of the methodology employed for our evaluation. Next, we validate the effectiveness of our proposed \textit{GPF\textsubscript{FAE}} metric on various procedural-fair and \mbox{-unfair} models, thereby demonstrating its capability to accurately distinguish procedural-fair and -unfair models.
Subsequently, we further investigate the relationship between \textit{GPF\textsubscript{FAE}} and the degree of procedural fairness, exploring how our metric captures and quantifies the level of procedural fairness or unfairness of an ML model. We also examine the impact of using different FAE methods on the evaluation results and computational efficiency of the \textit{GPF\textsubscript{FAE}} metric, verifying the reliability and robustness of the proposed metric. Furthermore, we explore the relationship between procedural fairness and distributive fairness of the ML model, revealing their interactions and potential implications. Additionally, we analyze the choice of the hyper-parameter $n$ in the \textit{GPF\textsubscript{FAE}} metric, considering its impact on the metric's performance and accuracy.
Finally, we critically discuss the limitations associated with our proposed metric, ensuring a comprehensive understanding of its capabilities and potential limitations, and provide promising solutions. The source code of this work is available at https://github.com/oddwang/GPF-FAE.

\subsubsection{Experimental Setup}

Here we describe the relevant settings in our experiments, including the dataset used, the model used, and the relevant parameter settings.

\begin{figure}[h]
    \centering
    \includegraphics[width=0.45\linewidth]{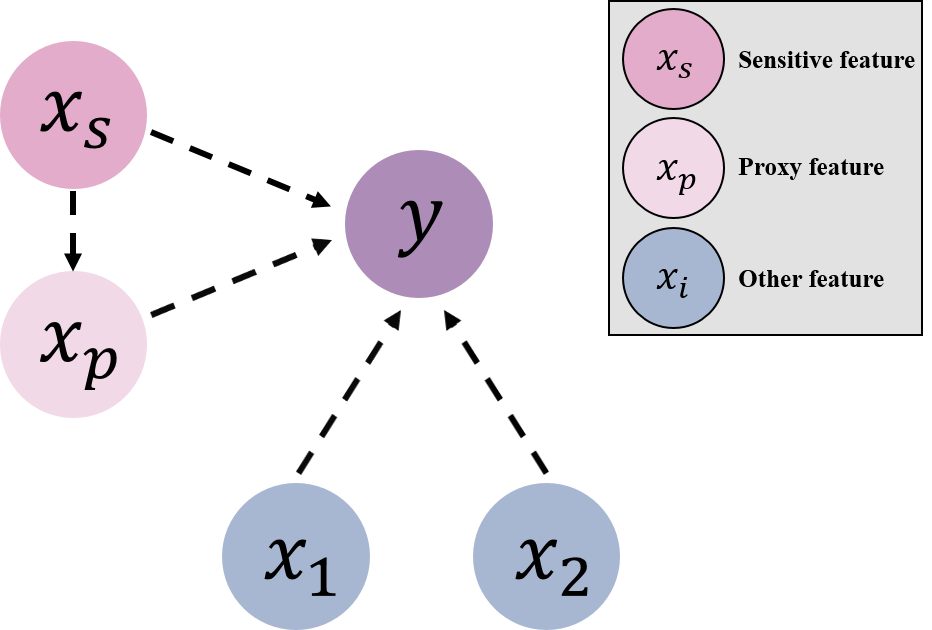}
    \caption{Relationship graph for the synthetic dataset. The dotted arrows indicate influence relationships, e.g., $x_1\dashrightarrow y$ means that the value of $x_1$ affects $y$.}
    \label{fig:synthetic_data}
\end{figure}

\paragraph{Datasets} In this paper, we conducted experiments on a synthetic dataset and eight real-world datasets that are widely used in fair ML~\cite{le2022survey}. For the synthetic dataset, we referred to the generation scheme of Jones \textit{et al.}~\citeyear{jones2020metrics}, which has \num[group-separator={,}]{10000} data points, including two non-sensitive features $x_1$ and $x_2$, one sensitive feature $x_s$, one proxy feature $x_{p}$ for sensitive feature $x_s$, and one label class $y$. Among them, $x_1\sim N(0,1)$, $x_2\sim N(0,1)$, while for the sensitive feature $x_s$, 6000 data points are set to 1 and 4000 data points are set to 0 (representing the advantaged and disadvantaged groups, respectively). The proxy feature $x_{p}\sim N (x_s, 0.1)$, and the label class $y=[\frac{1}{1+e^{-t}}+0.5]$, where $t=w_0+w_1x_1+w_2x_2+w_3x_{s}+w_4x_{p}+N(0,1)$. In this paper, $w_0$, $w_1$, $w_2$, $w_3$, and $w_4$ were taken as $-0.2$, 1.5, 0.5, 0.5, and 0.5, respectively. The relationship graph between features and class label in the synthetic dataset is shown in Fig. \ref{fig:synthetic_data}. In addition, eight real-world datasets widely used in ML fairness~\cite{le2022survey}, namely \textit{Adult}~\cite{Dua:2019}, \textit{Dutch}~\cite{van20002001}, \textit{LSAT}~\cite{wightman1998lsac}, \textit{COMPAS}~\cite{angwin2016machine}, \textit{German}~\cite{Dua:2019}, \textit{KDD}~\cite{Dua:2019}, \textit{Bank}~\cite{Dua:2019}, and \textit{Default}~\cite{Dua:2019}, are also used in our experimental study. Table \ref{tab:dataset} summarizes these datasets. Each dataset is preprocessed in the same way: label encoding of categorical features, and then normalizing all features by Z-score. In addition, each dataset was randomly divided into training and test sets with a ratio of 4:1, denoted as $\bm{\mathcal{D}}_{train}$ and $\bm{\mathcal{D}}_{test}$, respectively.

\begin{table}[t]
    \centering
    \caption{The eight real-world datasets and one synthetic dataset used in the study. “$|\bm{\mathcal{D}}|$'', “$|X|$'', “\textit{DP}'', and “$S$'' denote the number of data points, the number of features, the \textit{DP} value of the dataset itself, and the sensitive attribute under consideration, respectively.}
    \label{tab:dataset}
    \begin{tabular}{ccccccc}
        \toprule
            \multirow{1}{*}{Dataset} & \multirow{1}{*}{$|\bm{\mathcal{D}}|$} & \multirow{1}{*}{$|X|$} & \multirow{1}{*}{\textit{DP}} & \multirow{1}{*}{$S$} & Disadvantaged Group & Advantaged Group \\
        \midrule
        Adult & \num[group-separator={,}]{48842} & 14 &  0.195 & Sex & Female & Male \\
        Dutch & \num[group-separator={,}]{60420} & 11 & 0.298 & Sex & Female & Male \\
        LSAT & \num[group-separator={,}]{20798} & 11 & 0.198 & Race & Non-White & White \\
        COMAPS & \num[group-separator={,}]{6172} & 7 & 0.132 & Race & African-American & Others \\
        German & \num[group-separator={,}]{1000} & 20 & 0.075 & Sex & Female & Male \\
        KDD & \num[group-separator={,}]{284556} & 36 & 0.076 & Sex & Female & Male \\
        Bank & \num[group-separator={,}]{45211} & 16 & 0.048 & Marital &  Married & Single\\
        Default & \num[group-separator={,}]{30000} & 23 & 0.034 & Sex & Female & Male \\
        Synthetic & \num[group-separator={,}]{10000} & 4 & 0.199 & $x_s$ & 0 & 1 \\
        \bottomrule
    \end{tabular}
\end{table}

\paragraph{Parameter Setting} In our experiments, a two-layer artificial neural network (ANN) was trained with the ReLU activation function, Adam optimizer, and binary cross-entropy (BCE) loss function on each dataset. Each ANN model was fully connected to the hidden layer with 32 nodes, except for the \textit{German} and \textit{KDD} datasets, which had a high number of features, and we set the hidden layer with 64 nodes. The number of iterations and the learning rate were set to 300 and 0.01, respectively. 

The MMD significance level was set to 0.05, which means that there is a significant difference between the explanation distribution of the two groups when $\textit{GPF\textsubscript{FAE}}\leq 0.05$, i.e., the model is procedural-unfair, and conversely, the model is procedural-fair. In addition, in our experiments, we took $n=100$ (the selection of parameter $n$ was chosen based on our experimentation and will be discussed in Section \ref{sec:par_n}), i.e., we selected 100 pairs of similar data points from $\textit{\textbf{X}}_1$ and $\textit{\textbf{X}}_2$. Specifically, 50 data points were randomly selected from $\textit{\textbf{X}}_1$ of the test set, and then the 50 data points that are most similar to each of them were found from $\textit{\textbf{X}}_2$ of the test set and put them into $\textit{\textbf{X}}^{'}_{1}$ and $\textit{\textbf{X}}^{'}_{2}$, respectively. We repeated this process in reverse, and finally, there were 100 data points in each of $\textit{\textbf{X}}^{'}_{1}$ and $\textit{\textbf{X}}^{'}_{2}$ with the minimum distance between each pair of them. 

We used three distributive fairness metrics, \textit{DP}, \textit{EO}, and \textit{EOD}, implemented in the open-source Python algorithmic fairness toolkit AI Fairness 360 (AIF360)~\cite{bellamy2018ai} to explore the relationship between procedural and distributive fairness in ML. Referring to the criteria of AIF360~\cite{bellamy2018ai}, we used 0.10 as the threshold of the group distributive fairness metrics~\cite{stevens2020explainability}. That is, when the \textit{DP}, \textit{EO}, and \textit{EOD} metrics are below 0.10, the model is regarded as distributive-fair; otherwise, the model is considered as distributive-unfair. Each experiment case was performed 10 times in independent runs using different random number seeds.

\subsubsection{Evaluation on Procedural-Fair and Procedural-Unfair Models}
\label{sec:EPFM}

In this part, we evaluate the proposed metric \textit{GPF\textsubscript{FAE}} by constructing a procedural-fair model and a procedural-unfair model on each dataset, respectively, so that we can examine whether the proposed metric \textit{GPF\textsubscript{FAE}} accurately identifies whether an ML model is procedural-fair or procedural-unfair. 

We first specify how to construct procedural-fair and -unfair models, respectively. To obtain the procedural-fair model, we only used “fair features'' to train the model. The underlying assumption is consistent with Grgi{\'c}-Hla{\v{c}}a \textit{et al.}~\citeyear{grgic2018beyond} that the models trained with fair features are procedural-fair. On the synthetic dataset, we used two features $x_1$ and $x_2$ that are unrelated to the sensitive attribute. On each real-world dataset, since there is no “ground truth'' or definitive criteria to guide the selection of “fair features'', we calculated the correlation of each feature with the considered sensitive attribute using the Pearson correlation coefficient. We considered the features with a correlation coefficient below 0.10 as “fair features'' for training the model (except on the \textit{COMPAS} and \textit{LSAT} datasets, where we used 0.20 as the threshold due to the high correlation coefficient of each feature with the sensitive attribute).

\begin{figure}[htpb]
    \centering
    \includegraphics[width=0.6\linewidth]{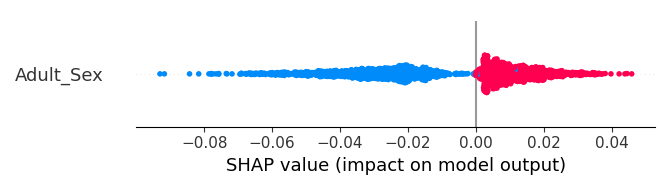}
    \caption{The “sex'' feature on the \textit{Adult} dataset is used as an example to illustrate the criteria for constructing a significantly procedural-unfair model. Each red and blue point represents a data point from the male and female groups, respectively.}
    \label{fig:Adult_sex}
\end{figure}

We ensured that the model is significantly procedural-unfair by examining the explanation results obtained by the FAE method on the sensitive attribute. Specifically, we ensured that the model is a procedural-unfair model when different values of the sensitive attribute, such as male and female, have opposite impacts on the model's decisions. Taking the \textit{Adult} dataset as an example, as shown in Fig. \ref{fig:Adult_sex}, the male (red points) and female (blue points) groups have positive and negative impacts on the final decision, respectively, indicating that the model is procedural-unfair in favor of the male group. However, it is important to note that this condition is sufficient but not necessary, as there may be proxy attributes at play that do not guarantee procedural fairness, even when different values of the sensitive attribute have the same impact on the decision.

We constructed procedural-unfair models on each dataset in two steps. First, for the four inherently fairer datasets, \textit{German}, \textit{KDD}, \textit{Bank}, and \textit{Default}, with \textit{DP} values less than 0.10 (as shown in Table \ref{tab:dataset}), we randomly resampled the data points labeled as “1'' in the advantaged group until their dataset achieves a \textit{DP} value greater than 0.10, and called them \textit{Unfair German}, \textit{Unfair KDD}, \textit{Unfair Bank}, and \textit{Unfair Default}. 

Second, on datasets except \textit{COMPAS} and \textit{LSAT}, we used BCE as the loss function to obtain a procedural-unfair model that meets the requirements. On the \textit{COMPAS} and \textit{LSAT} datasets, we used “$BCE+\lambda \times DP$'' ($\lambda$ is slightly less than 0) as the loss function to obtain the procedural-unfair model. This is because we were surprised to find that on both datasets, the models obtained by training directly with BCE as the loss function are not significant procedural-unfair models, as they do not satisfy the conditions required by our procedural-unfair model, although the datasets themselves are unfair (this case will be discussed further below). Eventually, the parameter $\lambda$ on the \textit{COMPAS} and \textit{LSAT} are set to -0.1 and -0.05, respectively. The selection criterion of parameter $\lambda$ was decided according to whether the FAE explanation results on the sensitive attribute of the trained model satisfy our requirements for the constructed procedural unfairness model. Obviously, the other normal parameter values that can satisfy the requirements are equally feasible. 

\begin{figure}[h]
    \centering
    \includegraphics[width=0.55\linewidth]{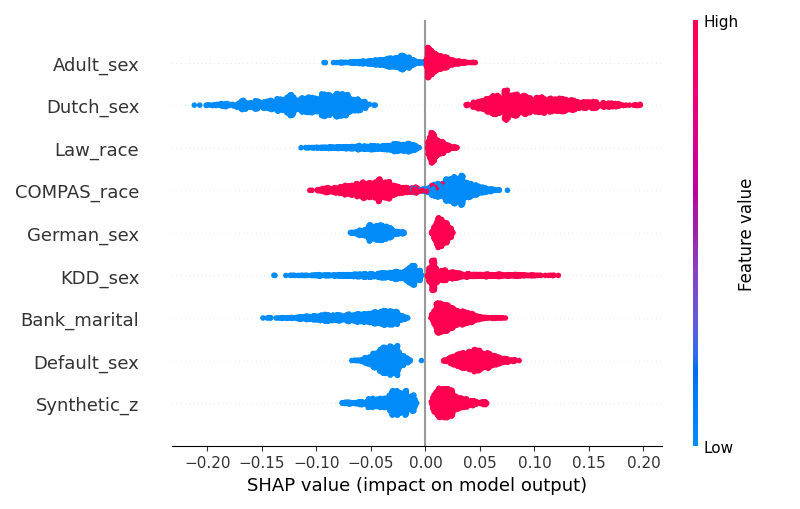}
    \caption{FAE explanation results obtained for the sensitive attribute on each constructed procedural-unfair model, where red and blue represent the advantaged and disadvantaged groups, respectively.}
    \label{fig:unfair_model}
\end{figure}

Finally, we present the FAE explanation results of the procedural-unfair model constructed on each dataset on the sensitive attribute, as shown in Fig. \ref{fig:unfair_model}. It demonstrates that the procedural-unfair models we constructed on the nine datasets all satisfy our requirement that they all significantly favor the advantaged group. It is worth noting that the distribution of explanations for the sensitive attribute in the \textit{COMPAS} dataset in Fig. \ref{fig:unfair_model} is the opposite of the other datasets. This is because the task of the \textit{COMPAS} dataset is to assess whether a criminal will re-offend, and contrary to the other tasks, predicting a positive category (i.e., will have recidivism) is instead disadvantageous, and thus the distribution of its explanations is exactly the opposite, which precisely means that the model favors the advantaged group.

The evaluation results of the \textit{GPF\textsubscript{FAE}} metric on the procedural-fair and -unfair models constructed on the nine datasets are shown in Tables \ref{tab:pf_model} and \ref{tab:puf_model}, respectively. We also present the evaluation results of the three distributive fairness metrics, \textit{DP}, \textit{EO}, and \textit{EOD}, to analyze the relationship between procedural fairness and distributive fairness afterward.

\begin{table}[h]
    \centering
    \caption{The evaluation results of \textit{GPF\textsubscript{FAE}}, \textit{DP}, \textit{EO}, and \textit{EOD} metrics on each dataset under the constructed procedural-fair models. Among them, \textit{DP}, \textit{EO}, and \textit{EOD} metrics considered to be distributive-unfair are underlined. The $\uparrow$ means that the larger the metric, the better, and $\downarrow$ means the opposite.}
    \label{tab:pf_model}
    \begin{tabular}{cccccc}
    \toprule
    Dataset & Feature Number & \textit{GPF\textsubscript{FAE}} $\uparrow$ & \textit{DP} $\downarrow$ & \textit{EO} $\downarrow$ & \textit{EOD} $\downarrow$ \\
    \midrule
    \multirow{1}{*}{Adult} & 10 & 0.943 & 0.079 & 0.032 & 0.019 \\
    \multirow{1}{*}{Dutch} & 9 & 1.000 & \underline{0.140} & \underline{0.102} & 0.051 \\
    \multirow{1}{*}{LSAT} & 5 & 0.990 & 0.003 & 0.002 & 0.003 \\
    \multirow{1}{*}{COMPAS} & 3 & 1.000 & \underline{0.182} & \underline{0.162} & \underline{0.149} \\
    \multirow{1}{*}{German} & 15 & 0.525 & 0.047 & 0.073 & 0.069 \\
    \multirow{1}{*}{KDD} & 17 & 1.000 & 0.018 & 0.023 & 0.012 \\
    \multirow{1}{*}{Bank} & 12 & 0.994 & 0.013 & 0.041 & 0.018 \\
    \multirow{1}{*}{Default} & 22 & 0.993 & 0.022 & 0.020 & 0.014 \\
    \multirow{1}{*}{Synthetic} & 2 & 1.000 & 0.015 & 0.095 & \underline{0.119} \\
    \bottomrule
    \end{tabular}
\end{table}

\begin{table}[h]
    \centering
    \caption{The evaluation results of \textit{GPF\textsubscript{FAE}}, \textit{DP}, \textit{EO}, and \textit{EOD} metrics on each dataset under the procedural-unfair models. Among them, \textit{DP}, \textit{EO}, and \textit{EOD} metrics considered to be distributive-fair are underlined. The $\uparrow$ means that the larger the metric, the better, and $\downarrow$ means the opposite.}
    \label{tab:puf_model}
    \begin{tabular}{cccccc}
    \toprule
    Dataset & Loss Function & \textit{GPF\textsubscript{FAE}}$\,\uparrow$ & \textit{DP}$\,\downarrow$ & \textit{EO}$\,\downarrow$ & \textit{EOD}$\,\downarrow$ \\
    \midrule
    \multirow{1}{*}{Adult} &  $BCE$ & 0.012 & 0.180 & \underline{0.099} & \underline{0.089}  \\
    \multirow{1}{*}{Dutch} & $BCE$ & 0.000 & 0.355 & 0.104 & 0.167  \\
    \multirow{1}{*}{LSAT} & $BCE-0.05 \times DP$ & 0.001 & 0.267 & 0.157 & 0.319 \\
    \multirow{1}{*}{COMPAS} & $BCE-0.1 \times DP$ & 0.001 & 0.367 & 0.365 & 0.326  \\
    \multirow{1}{*}{Unfair German} & $BCE$ & 0.000 & 0.117 & \underline{0.063} & \underline{0.093}  \\
    \multirow{1}{*}{Unfair KDD} & $BCE$ & 0.000 & 0.104 & 0.370 & 0.202  \\
    \multirow{1}{*}{Unfair Bank}& $BCE$ & 0.001 & 0.150 & 0.285 & 0.163  \\
    \multirow{1}{*}{Unfair Default} & $BCE$ & 0.000 & 0.126 & 0.144 & 0.100  \\
    \multirow{1}{*}{Synthetic}& $BCE$ & 0.000 & 0.251 & 0.111 & 0.126  \\
    \bottomrule
    \end{tabular}
\end{table}

Tables \ref{tab:pf_model} and \ref{tab:puf_model} show that our \textit{GPF\textsubscript{FAE}} metric accurately distinguishes between procedural-fair and -unfair models. On the constructed procedural-fair models, the \textit{GPF\textsubscript{FAE}} values are close to or at 1.0 on all datasets except on the \textit{German} dataset, which proves that \textit{GPF\textsubscript{FAE}} considers the distribution of the explanations of the two groups to be very consistent, i.e., the decision process of the model is very fair. In contrast, the \textit{GPF\textsubscript{FAE}} values on the \textit{German} dataset are relatively small and will be discussed further in Section \ref{sec:limitation}. On the other hand, on all the constructed procedural-unfair models, the \textit{GPF\textsubscript{FAE}} metric values are close to or reach 0.0, which means that the \textit{GPF\textsubscript{FAE}} metric considers the decision-making process of the two groups to be significantly different, i.e., it can accurately assess that the model's decisions are significantly unfair.

\begin{figure}[!htpb]
    \centering
    \subfigure[Disadvantaged group (African-American)]{
		\includegraphics[width=0.48\linewidth]{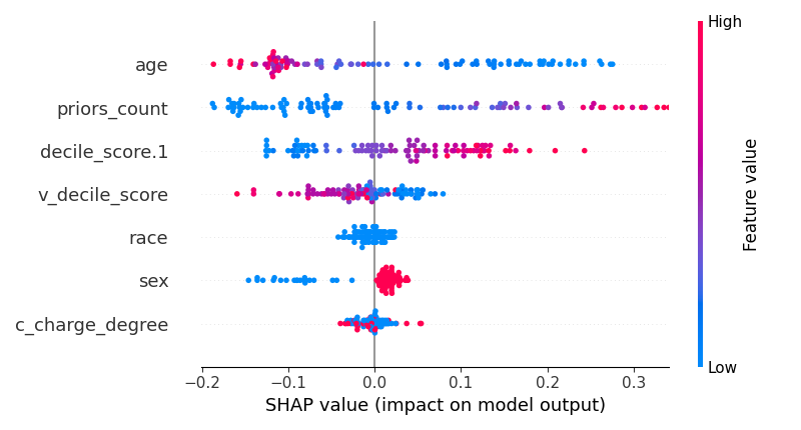}}
    \subfigure[Advantaged group (Others)]{
		\includegraphics[width=0.48\linewidth]{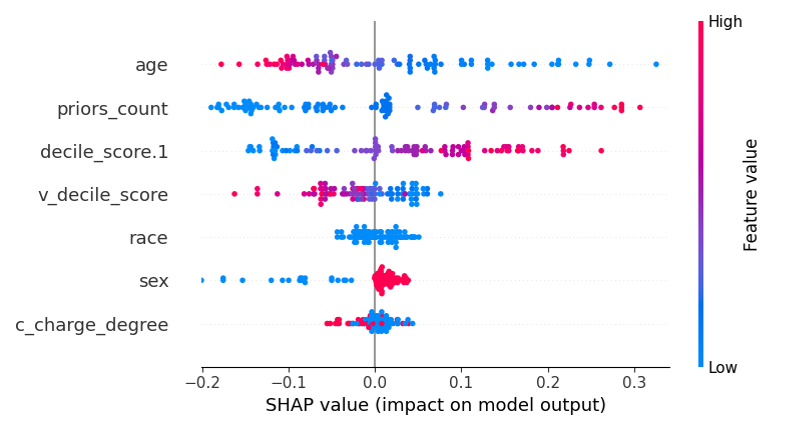}}
    \caption{Explanation results for disadvantaged and advantaged groups on the \textit{COMPAS} dataset.}
    \label{fig:COMAPS_explain}
\end{figure}

In addition, as previously noted, we performed a slight back-optimization of the \textit{DP} metric on the \textit{COMPAS} and \textit{LSAT} datasets to construct significant procedural-unfair models. This was due to the unexpected observation that the decision process of the models trained using BCE as the loss function on these two inherently unfair datasets did not exhibit a clear bias towards a particular group, especially on the \textit{COMPAS} dataset, as shown in Table \ref{tab:special_dataset}. To provide further insights, we present the explanation results of 100 data points each for the advantaged and disadvantaged groups on one of the independent runs on the \textit{COMPAS} dataset, as shown in Fig. \ref{fig:COMAPS_explain}. On that independent run, the \textit{GPF\textsubscript{FAE}} metric value is 0.690.

\begin{table}[h]
    \centering
    \caption{The evaluation results of \textit{GPF\textsubscript{FAE}}, \textit{DP}, \textit{EO}, and \textit{EOD} metrics on models obtained directly using BCE as a loss function for the \textit{COMPAS} and \textit{LSAT} datasets. The $\uparrow$ means that the larger the metric, the better, and $\downarrow$ means the opposite.}
    \label{tab:special_dataset}
    \begin{tabular}{cccccc}
    \toprule
    Dataset & Dataset \textit{DP} $\downarrow$ & \textit{GPF\textsubscript{FAE}} $\uparrow$ & \textit{DP} $\downarrow$ & \textit{EO} $\downarrow$ & \textit{EOD} $\downarrow$ \\
    \midrule
    COMPAS & 0.132 & 0.619 & 0.239 & 0.214 & 0.193 \\
    LSAT & 0.198 & 0.422 & 0.201 & 0.104 & 0.240 \\
    \bottomrule
    \end{tabular}
\end{table}

As we can see from Table \ref{tab:special_dataset}, although the inherent unfairness of the dataset and the model's distributive fairness metrics indicate significant unfairness, the model's decision process is relatively unbiased. Fig. \ref{fig:COMAPS_explain} also confirms the conclusion that the distribution of the explained results for the advantaged and disadvantaged groups is very similar, and it can be seen that there is no preference for a particular group on the sensitive attribute “race'', unlike the distribution of the sensitive attribute in the procedural-unfair model constructed on the \textit{COMPAS} dataset, which can be seen in Fig. \ref{fig:unfair_model}. This is also a counterexample to Grgi{\'c}-Hla{\v{c}}a \textit{et al.}~\citeyear{grgic2018beyond}'s definition of procedural fairness, showing that a model is modeled using intrinsically unfair features does not mean that it is a procedural-unfair model. Furthermore, this result is intriguing because the \textit{COMPAS} dataset has been used as a typical example to study fairness, and while it exhibits significant unfairness in terms of decision results, the decision process of the model obtained does not have a clear bias towards any certain group. In addition, for problems like the \textit{COMPAS} dataset that involve law and justice, we should pursue procedural fairness in addition to distributive fairness.

\subsubsection{Evaluating the Relationship between \textit{GPF\textsubscript{FAE}} Metric and the Degree of Group Procedural Fairness}
\label{sec:fairness_relationship}

So far, we have been evaluating our proposed metric \textit{GPF\textsubscript{FAE}} by constructing various procedural-fair and -unfair models on nine datasets. However, the preceding sections demonstrate more that \textit{GPF\textsubscript{FAE}} can correctly detect whether a model is procedural-fair or not rather than evaluate the degree of procedural fairness of the model. In this part, we aim to more thoroughly evaluate the relationship between \textit{GPF\textsubscript{FAE}} and procedural fairness, specifically by investigating whether the value of \textit{GPF\textsubscript{FAE}} can correctly reflect and assess the degree of procedural fairness of a model.

\begin{figure}[htpb]
    \centering
    \includegraphics[width=0.6\linewidth]{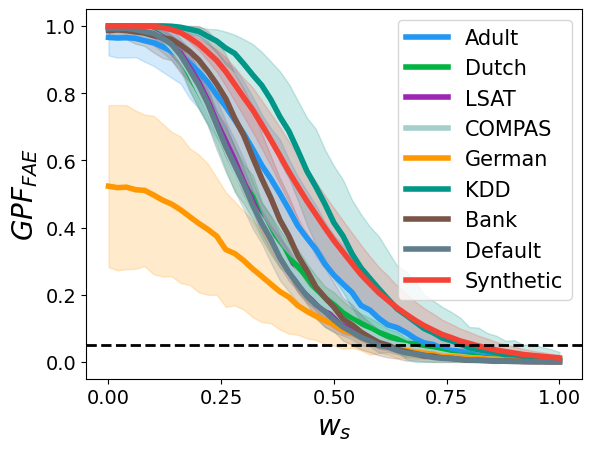}
    \caption{Trend of \textit{GPF\textsubscript{FAE}} metric values as $w_s$ increases. Figures show mean values over 10 random runs with a $1-\sigma$ error-bar.}
    \label{fig:relationship_GPF}
\end{figure}

However, it is difficult to construct various models with different degrees of procedural fairness/unfairness, especially on black-box models like ANNs. Therefore, instead of using ANNs, we use logistic regression (LR) models. On each dataset, we first constructed an LR model utilizing the features used to construct the procedural-fair model in Section \ref{sec:EPFM} and the sensitive attribute under consideration. Then, we manually control the weights $w_s$ of the LR model on the sensitive attribute, starting from 0.0 and increasing gradually. For the LR model, the weight $w_s$ determines the degree of influence of the sensitive attribute in the decision-making process. Obviously, as the value of $w_s$ gradually increases, the decision process of the model becomes increasingly unfair, and the trend of \textit{GPF\textsubscript{FAE}} is observed accordingly. The upper bounds of $w_s$ values are taken as 0.4, 0.8, 0.5, 0.2, 0.25, 0.35, 0.5, 0.3, and 5 for different datasets. We collected 50 parameters uniformly between 0.0 and the upper bounds to generate models with varying degrees of procedural unfairness. Taking the \textit{Adult} dataset as an example, we set $w_s$ to be 50 different parameters between [0.0, 0.4] one by one. Finally, we normalized the parameters $w_s$ on each dataset to facilitate the presentation of the experimental results, as shown in Fig. \ref{fig:relationship_GPF}.

As we can see in Fig. \ref{fig:relationship_GPF}, on each dataset, with the increase in the value of $w_s$, i.e., the unfairness degree of the model's decision process increases, the value of the \textit{GPF\textsubscript{FAE}} metric becomes progressively smaller. This illustrates that our proposed metric \textit{GPF\textsubscript{FAE}} is able to assess the degree of the model's procedural fairness correctly.

\subsubsection{Applicability of Different FAE Methods within the \textit{GPF\textsubscript{FAE}} Metric}

\begin{figure}[h]
    \centering
    \includegraphics[width=0.32\linewidth]{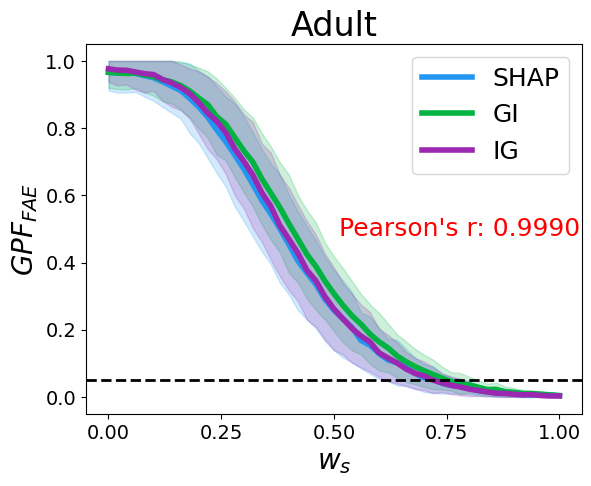}
    \includegraphics[width=0.32\linewidth]{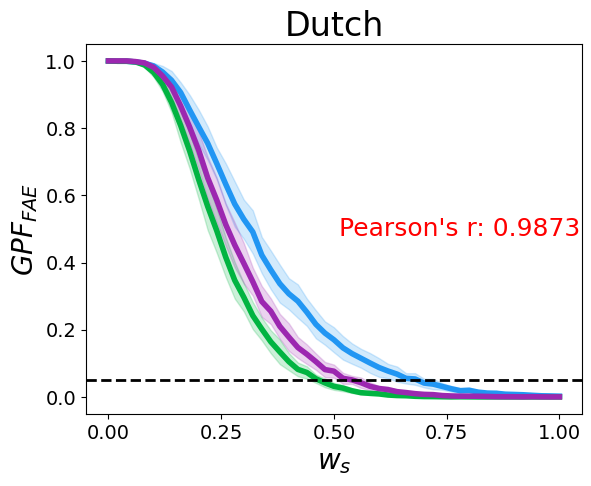}
    \includegraphics[width=0.32\linewidth]{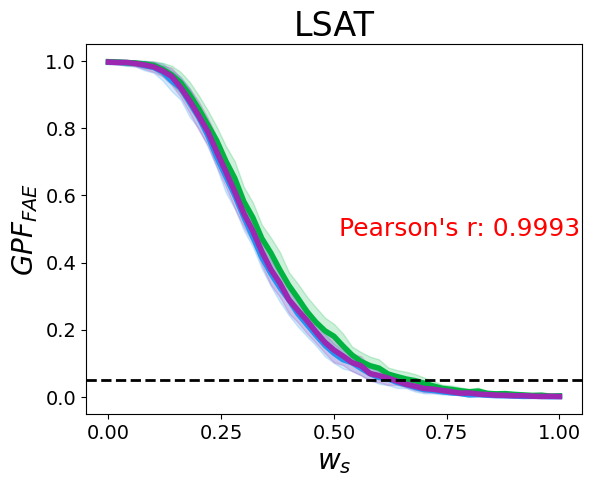}
    \includegraphics[width=0.32\linewidth]{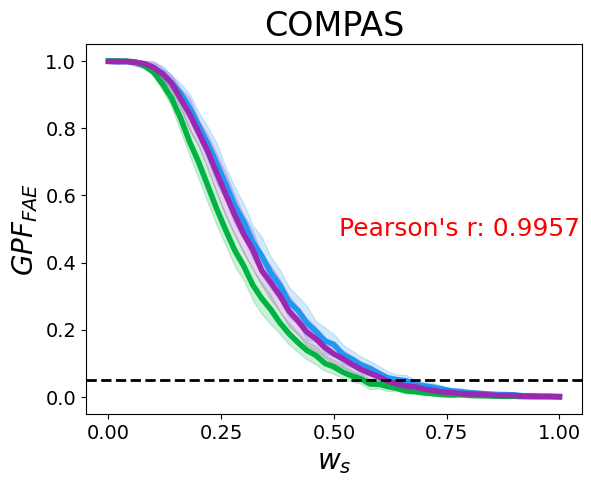}
    \includegraphics[width=0.32\linewidth]{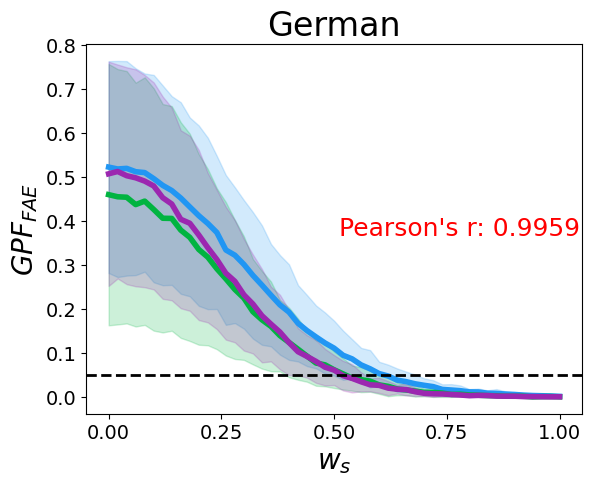}
    \includegraphics[width=0.32\linewidth]{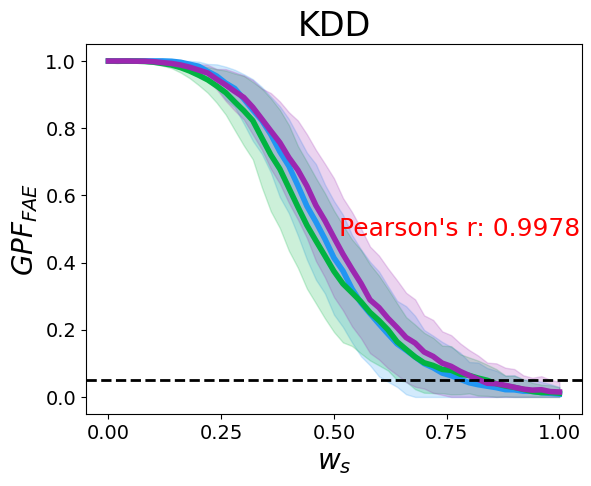}
    \includegraphics[width=0.32\linewidth]{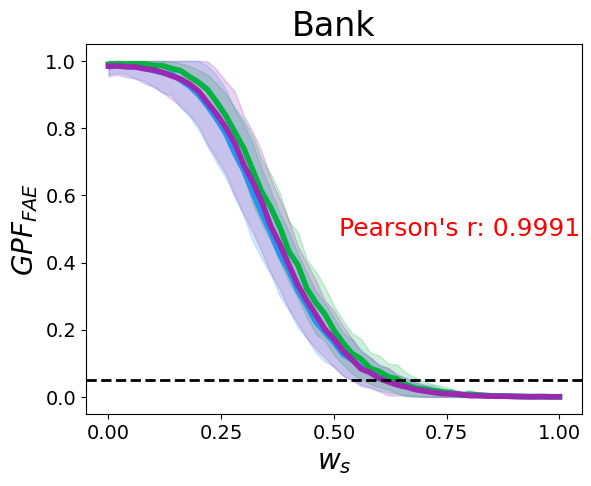}
    \includegraphics[width=0.32\linewidth]{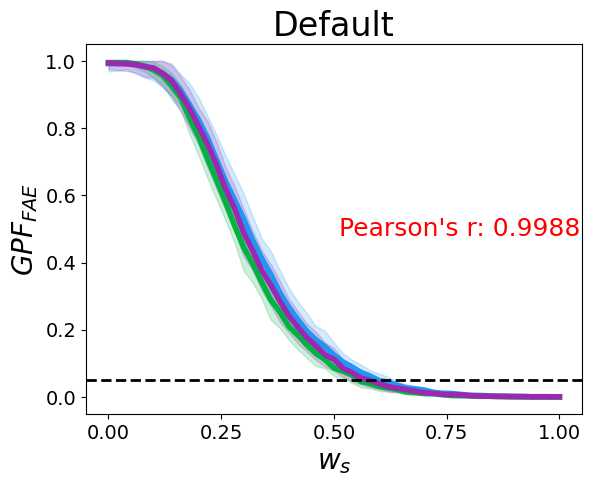}
    \includegraphics[width=0.32\linewidth]{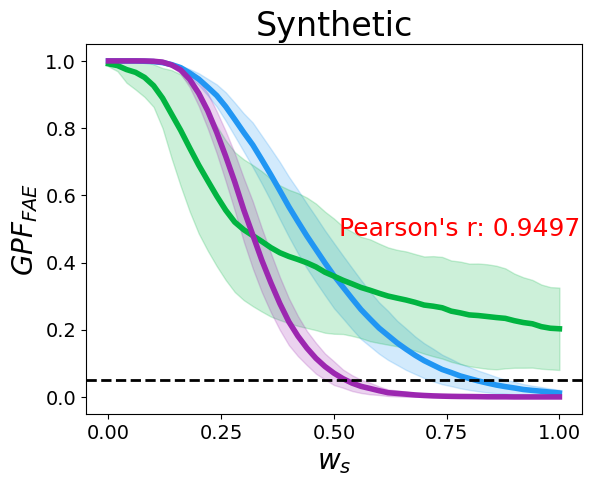}
        
    \caption{The trend of \textit{GPF\textsubscript{FAE}} metric values using different FAE methods as $w_s$ increases across various datasets. Figures show mean values over 10 random runs with a $1-\sigma$ error-bar.}
    \label{fig:relationship_GPF_dif_FAE}
\end{figure}

In this subsection, we explore whether FAE methods beyond SHAP can be employed to assess the procedural fairness of ML models using the proposed \textit{GPF\textsubscript{FAE}} metric. Specifically, we apply \textit{GPF\textsubscript{FAE}} with multiple FAE methods to LR models exhibiting varying degrees of procedural fairness constructed in Section~\ref{sec:fairness_relationship}. The considered FAE methods include the perturbation-based SHAP method~\cite{lundberg2017unified}, as well as two gradient-based methods: GI~\cite{shrikumar2016not} and IG~\cite{sundararajan2017axiomatic}. To quantify the consistency between these methods, we calculated the Pearson correlation coefficient between the \textit{GPF\textsubscript{FAE}} metric scores obtained by any two of the three methods and reported their average values. Their evaluation results are shown in Fig.~\ref{fig:relationship_GPF_dif_FAE}.

Fig.~\ref{fig:relationship_GPF_dif_FAE} illustrates that the results produced by the different FAE methods within the \textit{GPF\textsubscript{FAE}} framework are highly positively correlated, with Pearson correlation coefficients close to 1.0 across all datasets. This strong consistency suggests that the \textit{GPF\textsubscript{FAE}} metric is robust across various FAE methods and is not restricted to SHAP.

We also measured the runtime of \textit{GPF\textsubscript{FAE}} when evaluated on the constructed LR models using each FAE method, and the results are shown in Table~\ref{tab:run_time}. All experiments were conducted on a Linux server with an Intel(R) Xeon(R) Platinum 8358 CPU @ 2.60 GHz processor and 512 GB of memory. To ensure consistency, all experiments were executed using a single CPU core in single-threaded mode.

\begin{table}[htpb]
    \centering
    \caption{Average runtime (in seconds) of computing the \textit{GPF\textsubscript{FAE}} metric on the constructed LR model using different FAE method.}
    \label{tab:run_time}
    \begin{tabular}{l|ccc}
    \toprule
    Dataset & SHAP & GI & IG \\
    \midrule
    Adult & 128.84 & 0.22 & 0.25 \\
    Dutch & 69.11 & 0.18 & 0.15 \\
    LSAT & 9.36 & 0.13 & 0.15 \\
    COMPAS & 3.38 & 0.10 & 0.11 \\
    German & 134.07 & 0.15 & 0.11 \\
    KDD & 142.83 & 0.98 & 1.06 \\
    Bank & 127.03 & 0.18 & 0.17 \\
    Default & 145.44 & 0.15 & 0.14 \\
    Synthetic & 2.36 & 0.11 & 0.15 \\
    \midrule
    \midrule
    Mean & 84.71 & 0.24 & 0.25 \\
    \bottomrule 
    \end{tabular}
\end{table}

As can be seen from Table~\ref{tab:run_time}, the perturbation-based SHAP method incurs significantly higher computational cost compared to the gradient-based methods. Nonetheless, SHAP offers the advantage of being model-agnostic, making it suitable for explaining arbitrary ML models. In contrast, if the model under evaluation supports gradient computation, gradient-based FAE methods such as GI or IG can be employed to enhance evaluation efficiency significantly. For consistency, we continue to use SHAP as the default explanation method in the subsequent sections of this paper.

\subsubsection{Relationship between Procedural and Distributive Fairness}

In the following, we discuss the relationship between procedural and distributive fairness in the ML model. From Tables \ref{tab:pf_model} and \ref{tab:puf_model}, we can see that procedural and distributive fairness coincide in many cases, i.e., when the process is fair, the distributive fairness metrics are also fair, and vice versa. However, some procedural-fair models are considered as distributive-unfair and some procedural-unfair models are considered to be distributive-fair. Such a situation can be observed in Table \ref{tab:special_dataset} and Fig. \ref{fig:COMAPS_explain}, where the decision process is relatively fair, despite the significant unfairness in the three distributive fairness metrics. 

Thus, the relationship between procedural and distributive fairness in ML models is complex. They are sometimes coincidental with each other, but sometimes there are trade-offs and conflicts. Our findings here are consistent with the exploration of the relationship between the two in the humanities community~\cite{ambrose2013procedural}. For a more systematic exploration and in-depth analysis regarding the nuanced interactions and potential trade-offs between these two dimensions of fairness, one may refer to our follow-up work~\cite{wang2025procedural}.

\subsubsection{Study of the Parameter $n$ in the \textit{GPF\textsubscript{FAE}} Metric}
\label{sec:par_n}

\begin{figure}[htpb]
    \centering
    \subfigure[\textit{GPF\textsubscript{FAE}} under different $n$ on procedural fairness model]{
		\includegraphics[width=0.48\linewidth]{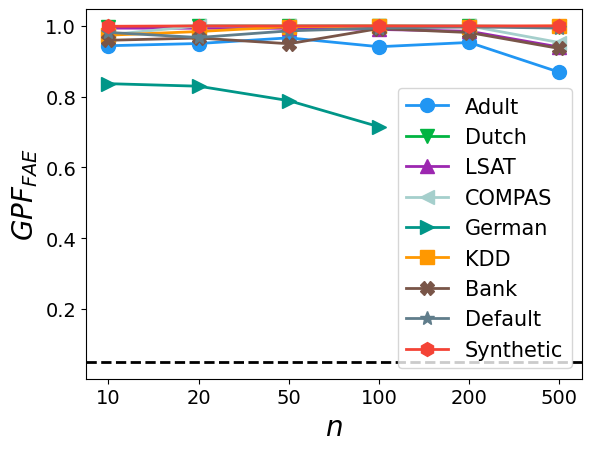}
    \label{fig:dif_n_a}
  }
    \subfigure[\textit{GPF\textsubscript{FAE}} under different $n$ on procedural unfairness model]{
		\includegraphics[width=0.48\linewidth]{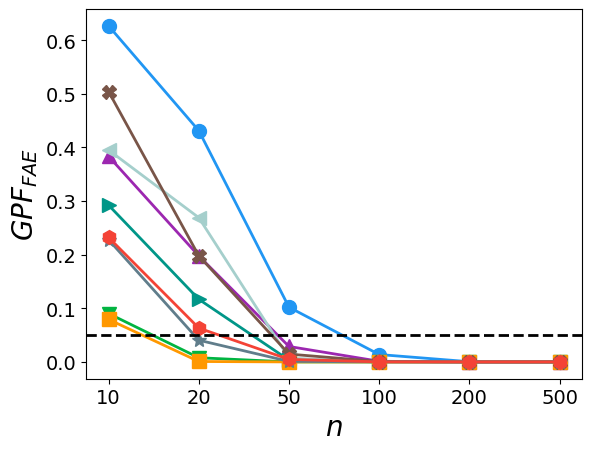}}
    \caption{The trend of \textit{GPF\textsubscript{FAE}} metric with parameter $n$ on the constructed procedural-fair and -unfair models.}
    \label{fig:dif_n}
\end{figure}

The parameter $n$ in the \textit{GPF\textsubscript{FAE}} metric determines how many pairs of similar data points are selected. In fact, ideally, we could compute the exact value of our proposed \textit{GPF\textsubscript{FAE}} metric on the entire dataset (i.e., $n=$ dataset size), but this would also result in longer computation times, so a trade-off is necessary. To choose the appropriate parameter $n$, we examined the impact of different values of parameter $n$. We took values of $n$ as 10, 20, 50, 100, 200, and 500, respectively, and observed the corresponding changes in the evaluation results of the \textit{GPF\textsubscript{FAE}} metric on the procedural-fair and -unfair models constructed in Section \ref{sec:EPFM}, as shown in Fig. \ref{fig:dif_n} (for the \textit{German} dataset, $n$ can only take a maximum of 100 due to dataset size limitations). The results indicate that for the procedural-fair models, the evaluation results remain relatively stable across different values of $n$. In contrast, for the procedural-unfair models, the \textit{GPF\textsubscript{FAE}} metric gradually detects the unfairness of the decision process of the model correctly and stabilizes as the parameter $n$ increases. However, it is worth noting that a larger value of $n$ implies more data points to be explained and a higher computational cost. In light of the trade-off between performance and efficiency, we choose $n=100$ in our paper.

\subsubsection{Limitations}
\label{sec:limitation}

Although our proposed metric \textit{GPF\textsubscript{FAE}} can accurately measure the procedural fairness of the model, it is limited by the requirement of obtaining $n$ pairs of similar data points with different values of the sensitive attribute. This may be challenging in situations where the dataset size is small or the data point distribution is very sparse. This can affect the final evaluation results. Intuitively, when the identified $n$ pairs of data points are not similar, their decision logics will naturally not be similar, and the \textit{GPF\textsubscript{FAE}} metric values will be small, even though the model is procedurally fair.

This phenomenon is especially noticeable in the \textit{German} dataset, which is the least data-rich dataset in the experiments of this paper, with only 1000 data, including 800 training data and 200 test data. Therefore, finding 100 pairs of similar data points in the 200 test data is obviously a challenge. As a result, the data points may not be similar enough to each other. The experimental results in Table \ref{tab:pf_model} show that the \textit{GPF\textsubscript{FAE}} metric values are close to or reach 1.0 for the constructed procedural-fair model evaluated on all other datasets, which means that the \textit{GPF\textsubscript{FAE}} metric considers the constructed model to be very procedural fairness. However, on the \textit{German} dataset, the \textit{GPF\textsubscript{FAE}} metric value is only 0.525. Similar observations can be made in other experiments, such as in Fig. \ref{fig:relationship_GPF} when exploring the relationship between the \textit{GPF\textsubscript{FAE}} metric and procedural fairness, where the \textit{GPF\textsubscript{FAE}} metric is close to 1.0 on all other datasets when $w_s$ is 0.0, but is less than 0.60 on the \textit{German} dataset with a very large variance. A similar phenomenon is also observed in Fig. \ref{fig:dif_n_a}, where the \textit{GPF\textsubscript{FAE}} metric value on the \textit{German} dataset is significantly smaller than the other datasets and is not stable.

To demonstrate that this is due to the small size of the dataset causing the selected $n$ pairs of data points are not similar, for the procedural-fair model constructed on the \textit{German} dataset, we compare two approaches of selecting similar data points: one using the test set only (i.e., $N=200$), and the other using the entire dataset (i.e., $N=1000$). The changes in the \textit{GPF\textsubscript{FAE}} metric and the average distance $\overline{d_{\textit{\textbf{x}}}}$ between the 100 pairs of similar data points are shown in Table \ref{tab:dif_data_size}.

\begin{table}[htpb]
    \centering
    \caption{On the $German$ dataset, the effect of selecting only similar data points from the test set ($N=200$) and from the entire data set ($N=1000$) on the \textit{GPF\textsubscript{FAE}} metric and the average distance $\overline{d_{\textit{\textbf{x}}}}$ between similar data points on the constructed procedural-fair model.}
    \label{tab:dif_data_size}
    \begin{tabular}{ccccc}
    \toprule
    Dataset & Feature Number & $N$ & \textit{GPF\textsubscript{FAE}} & $\overline{d_{\textit{\textbf{x}}}}$ \\
    \midrule
    \multirow{2}{*}{German} & \multirow{2}{*}{15} & 200 & 0.525 & 2.772 \\
    &  & 1000 & 0.708 & 2.331\\
    \bottomrule
    \end{tabular}
\end{table}

The results from Table \ref{tab:dif_data_size} demonstrate that the selected data point pairs are more similar to each other and the \textit{GPF\textsubscript{FAE}} metric values increase when selecting similar data points from the whole dataset instead of only the test set. This further proves that it is not that the constructed model is not fair enough, but that the data point size of the dataset is too small to find $n$ pairs of similar data points, which affects the final evaluation results.

\begin{figure}[htpb]
    \centering
    \subfigure{
		\includegraphics[width=0.6\linewidth]{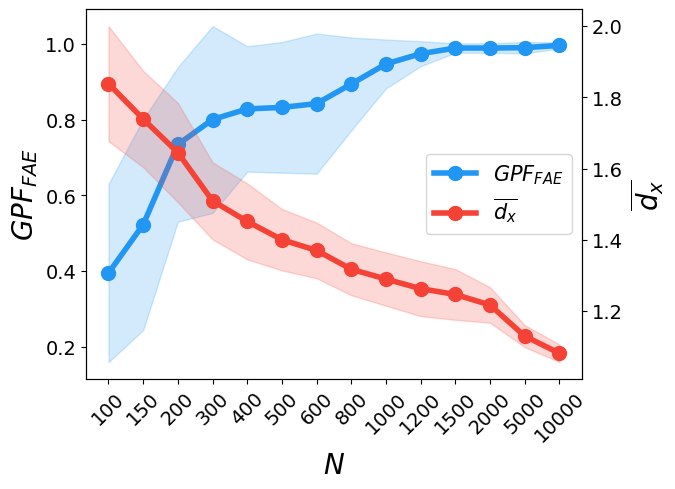}}
    \caption{The effect of selecting similar data points from different sizes of data $N$ on the average distance $\overline{d_{\textit{\textbf{x}}}}$ between similar data points and the \textit{GPF\textsubscript{FAE}} metric on the procedural-fair model constructed for the \textit{Dutch} dataset. Figures show mean values over 10 random runs with a $1-\sigma$ error-bar.}
    \label{fig:dif_N}
\end{figure}

To further explore the effect of the degree of similarity of the selected data points on the \textit{GPF\textsubscript{FAE}} metric, we observe the trend of the average distance $\overline{d_{\textit{\textbf{x}}}}$ between similar data points and the \textit{GPF\textsubscript{FAE}} metric on the constructed procedural-fair model on the \textit{Dutch} dataset by adjusting the size $N$ of the selected similar data point set. At different values of $N$, we all ensure a 50/50 data point size for the advantaged and disadvantaged groups. The \textit{Dutch} dataset is chosen because it has a sufficient number of data points, and the \textit{GPF\textsubscript{FAE}} metric value can reach 1.0 on the constructed procedural-fair model with a moderate number of features. The results are shown in Fig \ref{fig:dif_N}.

We can see that when the value of $N$ is small, the average distance $\overline{d_{\textit{\textbf{x}}}}$ is large and the value of the \textit{GPF\textsubscript{FAE}} metric is small. And as the value of $N$ increases, the average distance $\overline{d_{\textit{\textbf{x}}}}$ between the selected similar data point pairs steadily decreases, while the value of the \textit{GPF\textsubscript{FAE}} indicator increases steadily until it stabilizes at about $N=1200$. However, this value may vary with different datasets and dimensions. In general, when the size of the dataset is small, the evaluation results may not be accurate enough due to the difficulty of obtaining similar data points. Therefore, to ensure reliable assessment results, it is recommended to select similar data points from at least 1000 data points, thus ensuring sufficient similarity between the selected data point pairs. 

\subsubsection{Counterfactual Data Generation for Sparse Datasets}

In the previous subsection, we observed that the proposed \textit{GPF\textsubscript{FAE}} metric struggles to accurately assess procedural fairness when it is difficult to identify sufficiently similar data points, particularly in small-scale or sparse datasets. This limitation poses challenges to the metric's applicability in such scenarios. To address this issue, we propose a counterfactual data generation strategy: when direct matching fails due to data sparsity, we synthetically generate similar data points to serve as counterfactuals.

Specifically, given a data point $\textit{\textbf{x}}^{(i)}\in \bm{\mathcal{D}}_1$ to be matched, we first train a kernel density estimation (KDE) model on the other group $\bm{\mathcal{D}}_2$. Then, we sample $k$ candidate points $\{\tilde{\textit{\textbf{x}}}_j\}_{j=1}^{k} \sim \text{KDE}$, and select the generated point with the smallest Euclidean distance to $\textit{\textbf{x}}^{(i)}$ as its counterfactual $\textit{\textbf{x}}_{cf}^{(i)}$:

\begin{equation}
    \textit{\textbf{x}}_{cf}^{(i)}=\text{arg} \min_{\tilde{\textit{\textbf{x}}}_j\sim\text{KDE}}||\tilde{\textit{\textbf{x}}}_j-\textit{\textbf{x}}^{(i)}||_2.
\end{equation}

This process is repeated for all $n$ data points to be matched, ensuring that each is paired with the most similar candidate among the $k$ synthetically generated points generated by KDE in the opposite group. We evaluated this approach on the \textit{German} dataset using LR models with varying degrees of procedural fairness constructed in Section~\ref{sec:fairness_relationship}. Additionally, we investigated the effect of different values of $k$, ranging from 100 to 2000. The results are shown in Fig.~\ref{fig:relationship_german_KDE} below.

\begin{figure}[htpb]
    \centering
    \includegraphics[width=0.6\linewidth]{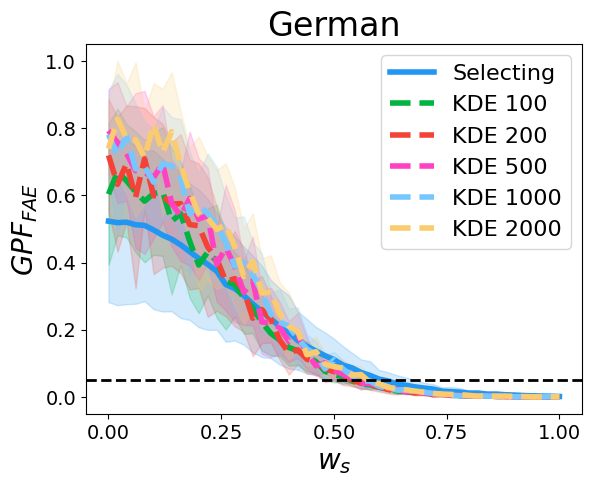}
    \caption{The trend of the \textit{GPF\textsubscript{FAE}} metric value as $w_s$ increases under different data point matching methods on the \textit{German} dataset. Figures show mean values over 10 random runs with a $1-\sigma$ error-bar.}
    \label{fig:relationship_german_KDE}
\end{figure}

As shown in Fig.~\ref{fig:relationship_german_KDE}, on the \textit{German} dataset, the counterfactual data generation strategy yields more accurate assessments of procedural fairness compared to direct matching with real data points. Additionally, increasing the number of generated candidates $k$ leads to modest improvements in evaluation accuracy, but the difference is not significant. These findings demonstrate that counterfactual data generation is a promising direction for extending the applicability of \textit{GPF\textsubscript{FAE}} to data-sparse environments.

Nonetheless, we emphasize that such generative methods should be viewed as a complementary strategy rather than a replacement for similarity-based matching. In practice, when the dataset is sufficiently large (e.g., more than 1000 samples, as discussed in Section~\ref{sec:limitation}), directly identifying real, similar data points remains the most reliable approach, as it avoids assumptions or potential artifacts introduced by synthetic generation. Future research could explore more advanced or task-specific counterfactual generation methods, such as those based on conditional generative models or diffusion processes, to further enhance evaluation reliability.

\section{Methods for Mitigating Procedural Unfairness in ML}
\label{sec:identify_improve}
In this section, we focus on mitigating procedural unfairness in ML models. Given a procedural-unfair ML model, we introduce a method to identify the sources that lead to its procedural unfairness, i.e., to find which features contribute to procedural unfairness (unfair features). By identifying these unfair features, we gain insights into the underlying causes of procedural unfairness. Then, we propose two distinct methods to mitigate the procedural fairness of the ML model based on the identified unfair features. The experimental results show that our proposed methods significantly improve the procedural fairness of models. 

\subsection{Identify Features that Lead to Procedural Unfairness}

In Section \ref{sec:metric}, we evaluate the procedural fairness of the model by assessing the degree of difference between the distributions of the two explanation sets $\textit{\textbf{E}}_1$ and $\textit{\textbf{E}}_2$. In this subsection, given a procedural-unfair ML model, we aim to identify the features that lead to its procedural unfairness, called unfair features (UFs). To detect these unfair features, we assess whether  the difference in each feature (that is, the importance score for each input feature) of the two explanation sets $\textit{\textbf{E}}_1$ and $\textit{\textbf{E}}_2$ is less than a predefined threshold, which is defined as
\begin{equation}
    UFs=\{i|d_{\Phi}(\textit{\textbf{E}}_{1,i},\textit{\textbf{E}}_{2,i})\leq \beta\};i=\{1,2,\dots,d\},
\end{equation}

\noindent where $\textit{\textbf{E}}_{1,i}$ and $\textit{\textbf{E}}_{2,i}$ denote the explanation results on the $i$-th feature in the explanation sets $\textit{\textbf{E}}_1$ and $\textit{\textbf{E}}_2$, respectively. $\beta$ is a threshold. Consistent with Section \ref{sec:metric}, we also used the MMD with an exponential kernel function as $d_{\Phi}$ to measure the difference between $\textit{\textbf{E}}_{1,i}$ and $\textit{\textbf{E}}_{2,i}$, and the $p$-value obtained by performing a permutation test on the generated kernel matrix was used as the final evaluation result. If the $p$-value is less than the threshold $\beta$, the feature is considered as a $UF$. Similar to the experiments in Section~\ref{sec:evaluate_metric}, the threshold $\beta$ was set to 0.05. However, users may set the threshold differently based on their specific requirements for the degree of procedural fairness.

In this subsection, we conduct experiments to identify $UFs$ on the nine procedural-unfair models constructed in Section~\ref{sec:EPFM} on the nine datasets. The average number of $UFs$ detected and the average runtime of the detection from 10 independent runs on each dataset are listed in Table~\ref{tab:unfair_number}.

\begin{table}[htpb]
    \centering
    \caption{The average number and standard deviation of $UFs$ detected and the average runtime} in 10 independent runs on each dataset.
    \label{tab:unfair_number}
    \begin{tabular}{cccc}
    \toprule
     Dataset & Loss Function & Number of $UFs$  & Runtime (in seconds)\\
     \midrule
     Adult & $BCE$ & 2.80 $\pm$ 0.92 & 254.26\\
     Dutch & $BCE$ & 2.10 $\pm$ 1.29 & 241.16\\
     LSAT & $BCE-0.05\times DP$ & 1.50 $\pm$ 0.97 & 242.46 \\
     COMPAS & $BCE-0.1\times DP$ & 1.40 $\pm$ 0.52 & 18.38 \\
     Unfair German & $BCE$ & 3.10 $\pm$ 1.37 & 262.64\\
     Unfair KDD & $BCE$ & 5.50 $\pm$ 2.72 & 279.74 \\
     Unfair Bank & $BCE$ & 1.40 $\pm$ 0.70 & 257.56 \\
     Unfair Default & $BCE$ & 1.00 $\pm$ 0.00 & 263.44 \\
     Synthetic & $BCE$ & 2.00 $\pm$ 0.00 & 4.02 \\
    \bottomrule
    \end{tabular}
\end{table}

According to our experiments, for the \textit{Synthetic} dataset, our method was able to accurately identify the sensitive attribute $x_s$ and its proxy attribute $x_p$ as $UFs$ in each independent run. On the \textit{Default} dataset, only the sensitive attribute “sex'' was detected as $UF$ each time, and this result is similar to the result of selecting fair features by Pearson correlation coefficient in Section \ref{sec:EPFM}, where \textit{Default} is the only dataset where the correlation coefficients of all other features and sensitive attributes are below the threshold 0.10. However, on most of the datasets, there is some fluctuation in the $UFs$ detected in different independent runs. For example, on the \textit{Adult} dataset, “sex'' and “relationship'' are the two unfair features detected each time, and it is correct that “sex'' is exactly the sensitive attribute under consideration, while “relationship'' includes about \num[group-separator={,}]{11000} data with the value of “husband'' or “wife'', which implies strong gender information, and can be regarded as a proxy attribute of “sex''. And features such as “hours-per-week'' and “race'' are sometimes seen as unfair features.

It is precisely because the features that cause procedural unfairness differ across models that our model-based approach is significant in identifying $UFs$. Our proposed approach can effectively identify $UFs$ that result in inconsistent decision-making processes between different groups and hence enable targeted interventions for improving fairness in ML models.

\subsection{Method 1: Retraining the Model by Eliminating $UFs$}

The first method to improve the procedural fairness of the model is straightforward, i.e., directly removing the detected $UFs$ and retraining the model. We evaluate the effectiveness of this method in three aspects: before and after removing the detected $UFs$ (1) the changes in the \textit{GPF\textsubscript{FAE}} metric to evaluate changes in procedural fairness; (2) the changes in the \textit{DP}, \textit{EO}, and \textit{EOD} metrics to evaluate changes in distributive fairness; and (3) the changes in model accuracy to evaluate changes in overall model performance.

\begin{figure*}[h]
    \centering
    \subfigure[\textit{GPF\textsubscript{FAE}} Metric ($\uparrow$)]{
        \label{fig:delete_a}
		\includegraphics[width=0.4\textwidth]{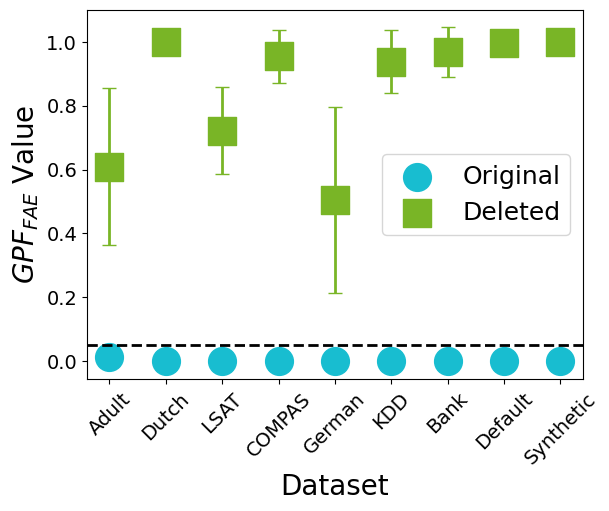}}
    \subfigure[\textit{DP} Metric ($\downarrow$)]{
		\includegraphics[width=0.4\textwidth]{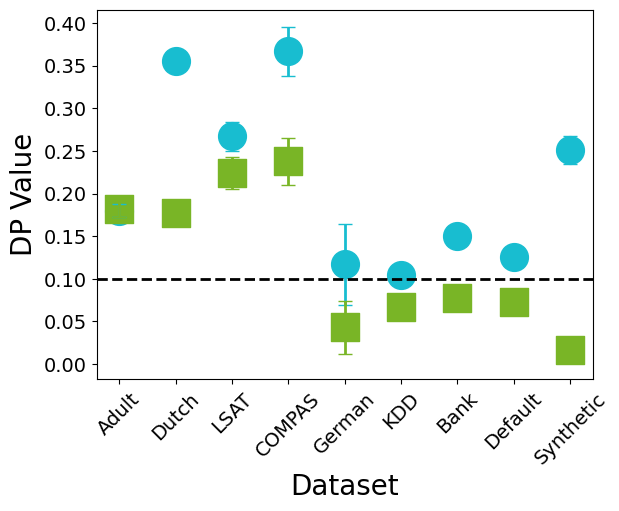}}
  \subfigure[\textit{EO} Metric ($\downarrow$)]{
		\includegraphics[width=0.4\textwidth]{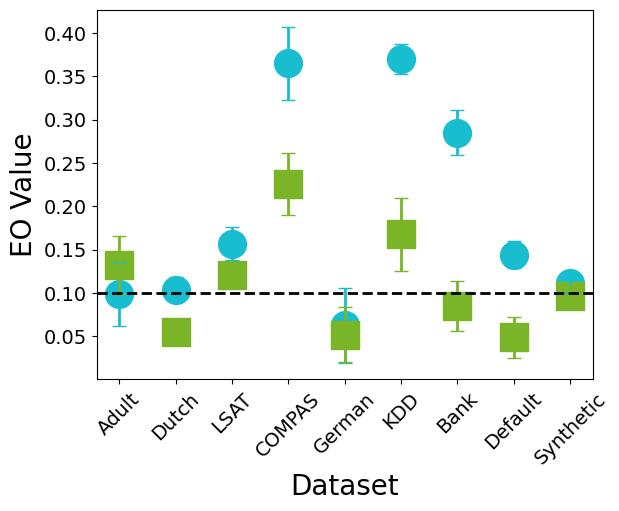}}
    \subfigure[\textit{EOD} Metric ($\downarrow$)]{
		\includegraphics[width=0.4\textwidth]{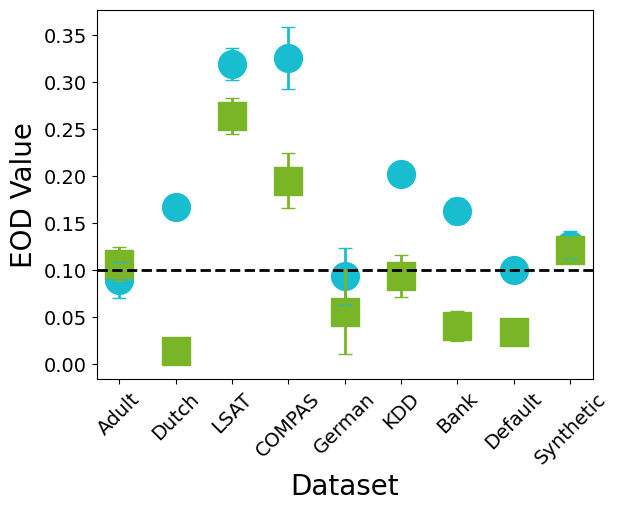}}
    \caption{Changes in \textit{GPF\textsubscript{FAE}}, \textit{DP}, \textit{EO}, and \textit{EOD} metrics before and after removing the detected $UFs$. Error bars indicate standard deviations. The $\uparrow$ means that the larger the metric, the better, and $\downarrow$ means the opposite.}
    \label{fig:delete_metric_change}
\end{figure*}

Fig. \ref{fig:delete_metric_change} presents the changes in \textit{GPF\textsubscript{FAE}}, \textit{DP}, \textit{EO}, and \textit{EOD} metrics of the model before and after removing the detected $UFs$. The results demonstrate that the procedural fairness of the model significantly improves after removing the detected $UFs$ on each dataset, especially on the \textit{Dutch}, \textit{COMPAS}, \textit{KDD}, \textit{Bank}, \textit{Default}, and \textit{Synthetic} datasets, where the \textit{GPF\textsubscript{FAE}} metric is able to reach or approach 1.0. In addition, there is also a considerable improvement in the distributive fairness metrics, with many of the datasets being able to fall below the distributive fairness threshold of 0.10, i.e., they can be considered distributive-fair models. This result not only validates the effectiveness of our proposed method in improving the fairness of the model but also demonstrates the accuracy of the detected $UFs$.

In addition, we show the changes in model accuracy on the test set before and after removing the detected $UFs$, as shown in Table \ref{tab:deleted_acc} below. We also include the average retraining time in the table. As can be seen, the removal of detected $UFs$ has the most significant impact on the accuracy of the model for the \textit{Synthetic} dataset. This is expected since two features were removed from the \textit{Synthetic} dataset with only four features in total, and the average accuracy decline over the nine datasets was not significant, at 0.8$\%$. Additionally, the retraining time cost is low, averaging 3.48 seconds.

\begin{table}[htpb]
    \centering
    \caption{Changes in model accuracy on the test set before and after removing the detected $UFs$ and the average retraining time for the model}. “$\Delta$'' is the difference between the two.
    \label{tab:deleted_acc}
    \begin{tabular}{l|cccc}
    \toprule
    \multirow{2}{*}{Dataset} & \multirow{2}{*}{Original Acc} & \multirow{2}{*}{Deleted Acc} & \multirow{2}{*}{$\Delta$} & Deleted Training Time\\
    & & & &  (in seconds) \\
    \midrule
    Adult & 85.1$\%\pm$0.3$\%$ & 84.9$\%\pm$0.3$\%$ & 0.2$\%\pm$0.2$\%$ & 3.76 \\
    Dutch & 83.7$\%\pm$0.4$\%$ & 82.0$\%\pm$0.5$\%$ & 1.7$\%\pm$0.3$\%$ & 4.42 \\
    LSAT & 89.9$\%\pm$0.3$\%$ & 89.9$\%\pm$0.3$\%$ & 0.0$\%\pm$0.1$\%$ & 2.44 \\
    COMPAS & 68.1$\%\pm$0.9$\%$ & 68.0$\%\pm$1.1$\%$ & 0.1$\%\pm$1.1$\%$ & 1.15 \\
    Unfair German & 79.7$\%\pm$2.6$\%$ & 78.0$\%\pm$1.8$\%$ & 1.7$\%\pm$1.2$\%$ & 1.76 \\
    Unfair KDD & 94.0$\%\pm$0.1$\%$ & 93.3$\%\pm$0.3$\%$ & 0.7$\%\pm$0.3$\%$ & 9.70 \\
    Unfair Bank & 86.2$\%\pm$0.4$\%$ & 86.1$\%\pm$0.3$\%$ & 0.1$\%\pm$0.3$\%$ & 3.93 \\
    Unfair Default & 79.7$\%\pm$0.4$\%$ & 79.5$\%\pm$0.3$\%$ & 0.2$\%\pm$0.2$\%$ & 2.82 \\
    Synthetic & 83.1$\%\pm$0.5$\%$ & 81.1$\%\pm$0.7$\%$ & 2.0$\%\pm$0.5$\%$ & 1.35\\
    \midrule
    \midrule
    Mean & 83.3$\%\pm$0.7$\%$ & 82.5$\%\pm$0.6$\%$ & 0.8$\%\pm$0.5$\%$ & 3.48\\
    \bottomrule 
    \end{tabular}
\end{table}

Overall, by directly removing the detected $UFs$, retraining the model is able to obtain a procedural-fair model, and the distributive fairness of the model is also improved, while the accuracy decrease of the model is only about 0.8$\%$ on average.

\subsection{Method 2: Modifying the Model by Reducing the Impacts of $UFs$}
\label{sec:improve_modify}

Although the method of directly removing the detected $UFs$ is simple and straightforward, the biggest problem is that it requires retraining the whole model. This may result in the decision logic of the retrained model differing significantly from the original model, i.e., not faithful to the original model, which may be undesirable. 

Rather than retraining the whole model, we prefer to improve fairness by adapting to the existing model, which typically allows its decision logic to be more faithful to the original model. Among them, Dimanov \textit{et al.} \citeyear{dimanoyouv2020you} proposed a method for modifying the existing model. For an already trained model $f_{\bm{\theta}}$ parameterized by $\bm{\theta}$ (in this paper, it is equivalent to $f$ in the previous section), they find a modified model $f'_{\bm{\theta}}$ with the following two properties:

\begin{enumerate}
    \item[1.] \textit{Model similarity}: the model has similar performance before and after modifying
    \begin{equation}
        \forall i,\, f'_{\bm{\theta}}(\textit{\textbf{x}}^{(i)})\approx f_{\bm{\theta}}(\textit{\textbf{x}}^{(i)}).
    \end{equation}
    \item[2.] \textit{Low sensitive attribute importance score}: the importance score provided by the explanation function $g$ on the sensitive attribute $s$ ($j$-th feature) decreases significantly after modifying
    \begin{equation}
        \forall i,\, |g(f'_{\bm{\theta}},\textit{\textbf{x}}^{(i)})_j|\ll |g(f_{\bm{\theta}},\textit{\textbf{x}}^{(i)})_j|.
    \end{equation}
\end{enumerate}

Straightforwardly, Dimanov \textit{et al.}~\citeyear{dimanoyouv2020you} reduce the impact of the sensitive attribute on the decision-making by modifying the model, while attempting to maintain the model's performance. 

However, the purpose of Dimanov \textit{et al.}~\citeyear{dimanoyouv2020you} is not to improve the fairness of the model. Instead, they found that after modifying the model, although the FAE obtained low importance scores for the sensitive attribute (tending towards 0.0), it was still a (distributive) unfairness model by metrics such as \textit{DP}, \textit{EO}, and \textit{EOD}. This leads the authors to conclude that the (distributive) fairness of the model cannot be evaluated by the importance score of the sensitive attribute obtained by FAE.

On the one hand, due to the presence of proxy attributes, we agree that it is difficult to assess the fairness of a model by evaluating the importance score of the sensitive attribute alone. However, we believe that this approach contributes to the procedural fairness of the model because of its ability to effectively reduce (or even eliminate) the influence of the sensitive attribute on the decision-making process.

However, as mentioned before, it is not enough to deal with just the sensitive attribute because of the presence of the proxy attribute. Therefore, in this paper, we present an improved method to modify the existing model by building upon the approach proposed by Dimanov \textit{et al.}~\citeyear{dimanoyouv2020you}. Instead of only modifying the sensitive attribute, we modified all the detected $UFs$, thereby reducing the impact of all the detected $UFs$ on the model decisions, and ultimately improving the procedural fairness of the model. Specifically, for the trained model $f_{\bm{\theta}}$, we modified it by optimizing an additional penalty term, called the explanation loss $\zeta$, weighted by the hyper-parameter $\alpha$, and normalized over all $m$ training data points:
\begin{equation}
\begin{split}
    \mathcal{L}'&=\mathcal{L}+\alpha\times \zeta, \\
    \zeta&=\sum_{k=1}^{|UFs|} \frac{1}{m}\times L^p([|\frac{\partial \mathcal{L}}{x^{(1)}_{k}}|,|\frac{\partial \mathcal{L}}{x^{(2)}_{k}}|,...,|\frac{\partial \mathcal{L}}{x^{(m)}_{k}}|]),  
\end{split}
\end{equation}

\noindent where $\mathcal{L}$ is the cross-entropy loss of the model, and we used $L^1$ norm to be consistent with Dimanov \textit{et al.}~\citeyear{dimanoyouv2020you}. The pseudo-code is shown in Algorithm \ref{alg:perturb_model}, where we took the number of iterations $\tau=200$, which is sufficient for convergence. In our experiments, we took explanation loss weight $\alpha=15$ (the selection of parameter $\alpha$ was chosen based on our experimentation and will be discussed later).

\begin{algorithm}
	\caption{Modifying the existing model to improve procedural fairness.} 
    \label{alg:perturb_model}
	\begin{algorithmic}[1]
        \Require Original trained model $f_{\bm{\theta}}$, unfair features $UFs$, input matrix $X\in\mathbb{R}^{m\times d}$ with corresponding labels $Y\in\mathbb{R}^{m\times 1}$, iteration number $\tau$, and explanation loss weight $\alpha$
        \Ensure The modified model $f'_{\bm{\theta}}$
        \State $i$ = 0
        \While{$i<\tau$}
            \State Calculate the cross entropy loss $\mathcal{L}$ with respect to $f_{\bm{\theta}}$
            \State Calculate the explanation loss $\zeta$ 
            \vspace{0.5\baselineskip}
            \Statex \centering $\zeta=\sum_{k=1}^{|UFs|} \frac{1}{m}\times L^p([|\frac{\partial \mathcal{L}}{x^{(1)}_{k}}|,|\frac{\partial \mathcal{L}}{x^{(2)}_{k}}|,...,|\frac{\partial \mathcal{L}}{x^{(m)}_{k}}|])$
            \vspace{0.5\baselineskip}
            \State \raggedright Calculate the total loss $\mathcal{L}'=\mathcal{L}+\alpha\times \zeta$
            \State Update model parameters with $\bigtriangledown_{\bm{\theta}} \mathcal{L}'$ using Adam
            \State $i = i + 1$
        \EndWhile 
        \State \Return the modified model $f_{\bm{\theta}}$ as $f'_{\bm{\theta}}$
	\end{algorithmic} 
\end{algorithm}

Straightforwardly, we improved the procedural fairness of the model by reducing the impact of all the detected $UFs$ on the model decisions. Similarly, we evaluated the performance of the method in terms of three aspects of the model's procedural fairness, distributive fairness, and changes in model performance before and after modification.

\begin{figure*}[h]
    \centering
    \subfigure[\textit{GPF\textsubscript{FAE}} Metric ($\uparrow$)]{
        \label{fig:perturb_a}
		\includegraphics[width=0.4\textwidth]{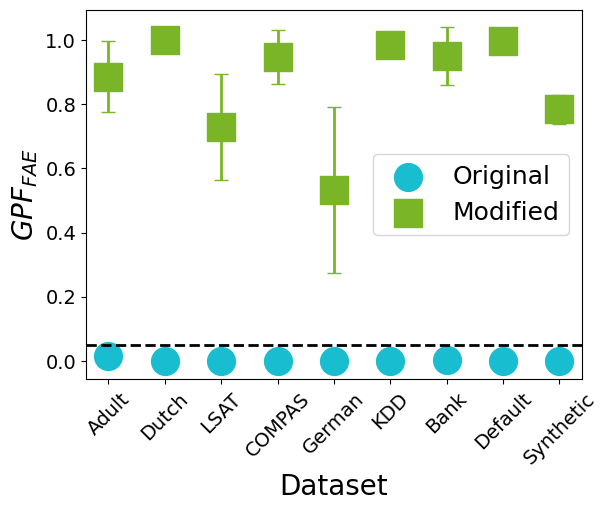}}
    \subfigure[\textit{DP} Metric ($\downarrow$)]{
		\includegraphics[width=0.4\textwidth]{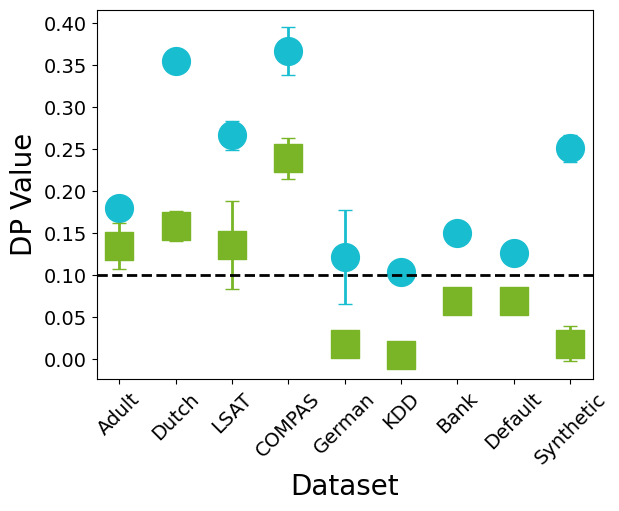}}
  \subfigure[\textit{EO} Metric ($\downarrow$)]{
		\includegraphics[width=0.4\textwidth]{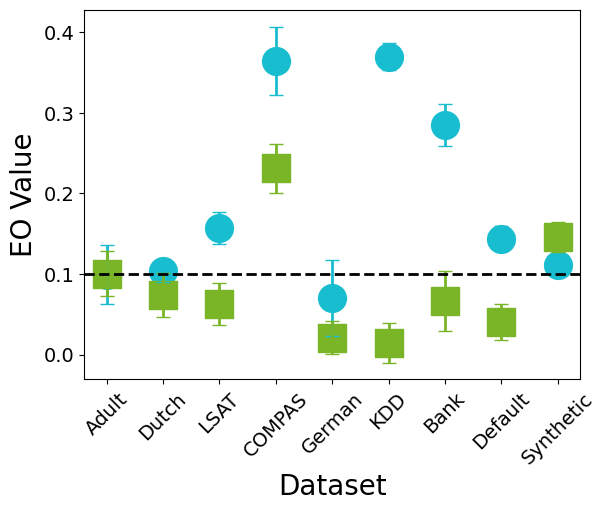}}
    \subfigure[\textit{EOD} Metric ($\downarrow$)]{
		\includegraphics[width=0.4\textwidth]{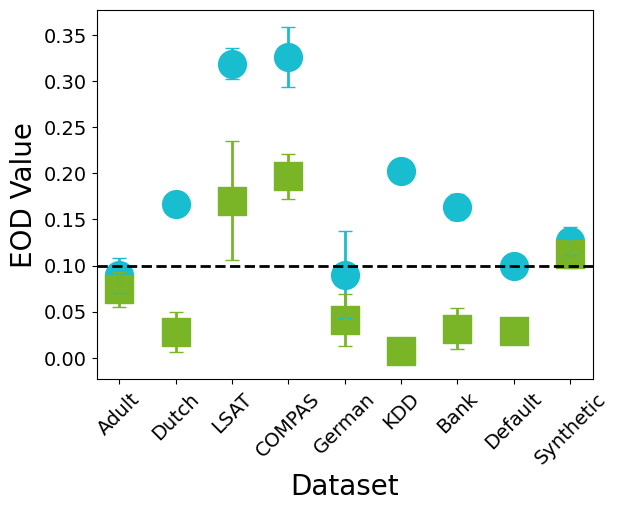}}
    \caption{Changes in \textit{GPF\textsubscript{FAE}}, \textit{DP}, \textit{EO}, and \textit{EOD} metrics before and after modifying the detected $UFs$. Error bars indicate standard deviations. The $\uparrow$ means that the larger the metric, the better, and $\downarrow$ means the opposite.}
    \label{fig:perturb_metric_change}
\end{figure*}

The changes in \textit{GPF\textsubscript{FAE}}, \textit{DP}, \textit{EO}, and \textit{EOD} metrics before and after modifying the model are shown in Fig. \ref{fig:perturb_metric_change}. We can see that the method significantly improves the procedural fairness of the model, while the distributive fairness of the model is also improved to some extent.

The comparison of model accuracy before and after modifying the existing model is shown in Table \ref{tab:perturb_acc}. The table also lists the average time required to modify the model. It illustrates that the modification method has a greater impact on model performance compared to the retraining method, with an average model accuracy loss of 1.8$\%$ (1.3$\%$ on the eight real-world datasets). In addition, it runs longer, but is still fast, averaging 24.97 seconds.

\begin{table}[h]
    \centering
    \caption{Changes in model accuracy on the test set before and after modifying the detected $UFs$ and the average time for modifying the model. “$\Delta$'' is the difference between the two.}
    \label{tab:perturb_acc}
    \begin{tabular}{l|cccc}
    \toprule
    \multirow{2}{*}{Dataset} & \multirow{2}{*}{Original Acc} & \multirow{2}{*}{Perturbed Acc} & \multirow{2}{*}{$\Delta$} & Perturbed Training Time\\
    & & & & (in seconds) \\ 
    \midrule
    Adult & 85.1$\%\pm$0.2$\%$ & 84.0$\%\pm$0.7$\%$ & 1.1$\%\pm$0.5$\%$ & 30.11 \\
    Dutch & 83.7$\%\pm$0.4$\%$ & 80.7$\%\pm$0.6$\%$ & 3.0$\%\pm$0.6$\%$ & 41.17 \\
    LSAT & 89.9$\%\pm$0.3$\%$ & 89.9$\%\pm$0.4$\%$ & 0.0$\%\pm$0.3$\%$ & 13.90 \\
    COMPAS & 68.1$\%\pm$0.9$\%$ & 67.9$\%\pm$1.5$\%$ & 0.3$\%\pm$1.2$\%$ & 1.31 \\
    Unfair German & 79.7$\%\pm$2.6$\%$ & 76.7$\%\pm$3.0$\%$ & 3.0$\%\pm$2.3$\%$ & 1.51 \\
    Unfair KDD & 94.0$\%\pm$0.1$\%$ & 91.9$\%\pm$0.4$\%$ & 2.1$\%\pm$0.5$\%$ & 89.73 \\
    Unfair Bank & 86.2$\%\pm$0.4$\%$ & 85.4$\%\pm$0.7$\%$ & 0.8$\%\pm$0.6$\%$ & 24.58 \\
    Unfair Default & 79.7$\%\pm$0.4$\%$ & 79.5$\%\pm$0.4$\%$ & 0.2$\%\pm$0.1$\%$ & 21.20 \\
    Synthetic & 83.1$\%\pm$0.5$\%$ & 77.6$\%\pm$0.8$\%$ & 5.5$\%\pm$0.7$\%$ & 1.25 \\
    \midrule
    \midrule
    Mean & 83.3$\%\pm$0.6$\%$ & 81.5$\%\pm$0.9$\%$ & 1.8$\%\pm$0.8$\%$ & 24.97\\
    \bottomrule 
    \end{tabular}
\end{table}

\begin{figure}[htpb]
    \centering
    \subfigure{
		\includegraphics[width=0.48\linewidth]{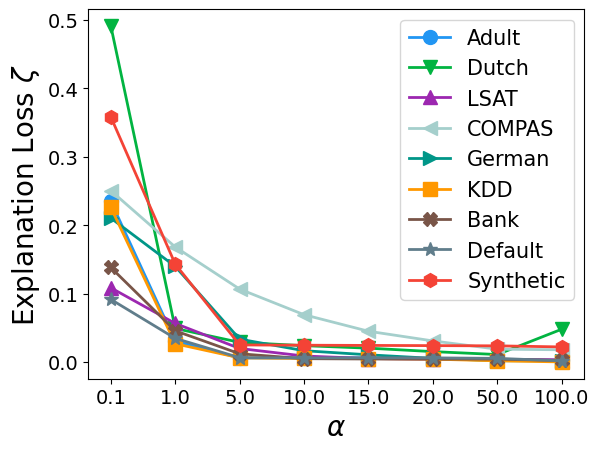}}
    \subfigure{
		\includegraphics[width=0.48\linewidth]{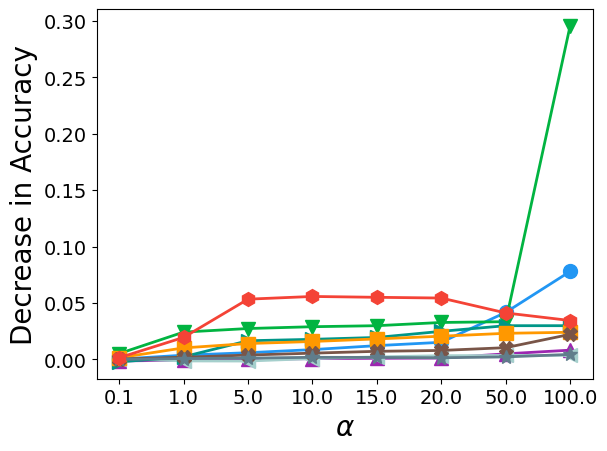}}
    \caption{Trends in the explanation loss $\zeta$ and decrease in accuracy of the modified model with different explanation loss weights $\alpha$.}
    \label{fig:dif_alpha}
\end{figure}

Finally, we discuss the impact of different explanation loss weights $\alpha$. We investigated the trends of the explanation loss $\zeta$ and decrease in accuracy of the modified model under different explanation loss weights $\alpha$ (taken as 0.1, 1, 5, 10, 15, 20, 50, 100, respectively), as shown in Fig. \ref{fig:dif_alpha}. The results indicate that, on the one hand, when the value of $\alpha$ is small, the explanation loss $\zeta$ is still large despite the low decrease of the model accuracy, i.e., it does not reduce the influence of $UFs$ on decision making; on the other hand, when $\alpha$ takes a large value, it has a disastrous effect on the performance of the model. When $\alpha$ takes values between [10, 20], the results exhibit good stability, and we can achieve a better trade-off between these two goals, i.e., we can reduce the explanation loss $\zeta$ while affecting the model performance in a smaller way. In the experiments of this paper, we set $\alpha=15$.

Fig. \ref{fig:dif_alpha} also demonstrates that there is a conflict between model performance and procedural fairness, and that users have the flexibility to control the $\alpha$ value to make a trade-off between model performance and procedural fairness.

\subsection{Similarities and Differences between the Two Methods}

The two approaches we proposed to improve the procedural fairness of the model are very similar. They both work by reducing or removing detected $UFs$ so that they no longer influence the model's decisions. They are both able to improve the procedural fairness of the model to a large extent, while the distributive fairness of the model is also improved.

\begin{figure*}[htpb]
    \centering
    \subfigure[Original Model]{
		\includegraphics[width=0.31\textwidth]{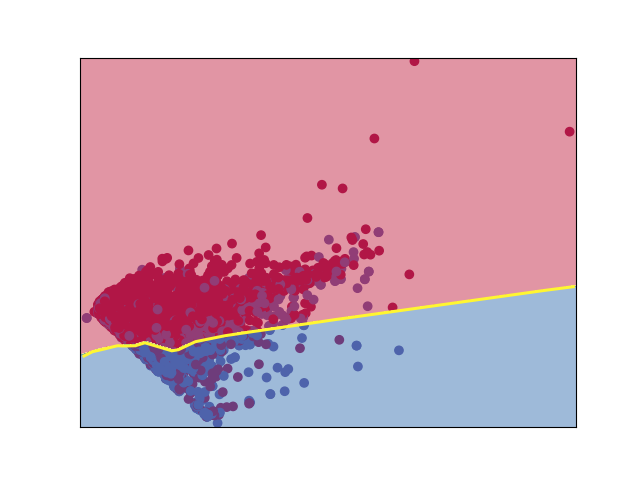}}
    \subfigure[Modified Model]{
		\includegraphics[width=0.31\textwidth]{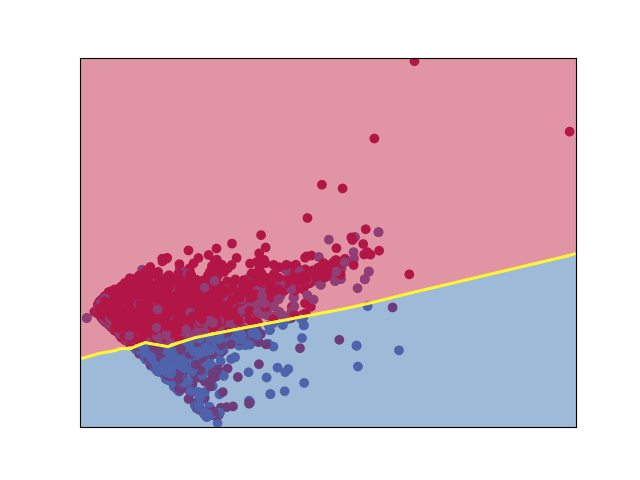}}
  \subfigure[Retrained Model]{
		\includegraphics[width=0.31\textwidth]{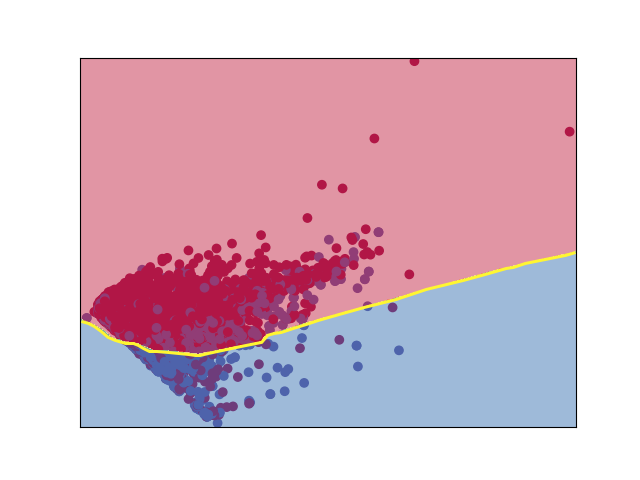}}
    \caption{Decision boundaries for the original, modified and retrained models visualized on the \textit{Unfair Default} dataset in 2D reduced input space (dimensionality reduction by PCA). The red and blue backgrounds indicate negative and positive predictions, respectively. Each point represents a 2D projection of each data point in the dataset, and its color indicates its true label.}
    \label{fig:decision_bound}
\end{figure*}

The primary difference between the two is that, compared to modifying the existing model, retraining after removing the $UFs$ tends to be less detrimental to the model's performance. However, since the modification approach is modified from the obtained model, the decision logic tends to be more “faithful'' to the original model, i.e., the decision logic of the two is more similar. To visualize this, we projected the decision boundaries of the original model, the modified model, and the retrained model on the \textit{Unfair Default} dataset in 2D PCA projected space, as shown in Fig. \ref{fig:decision_bound}. We can see that the left side of the decision boundary of the retrained model has changed significantly compared to the original model, while the modified model is very similar to the original model. In addition, compared to the retraining approach, the modification approach allows for more flexible and fine-grained trade-offs between model performance and procedural fairness by controlling the parameter $\alpha$.

\section{Conclusion}
\label{sec:conclusion}

In this paper, we first proposed a definition of procedural fairness in ML. Then, we introduce a novel group procedural fairness metric based on the FAE approach, called \textit{GPF\textsubscript{FAE}}. Through comprehensive experimentation on nine datasets, we demonstrated the efficacy of \textit{GPF\textsubscript{FAE}} in accurately assessing the procedural fairness of the ML model. After assessing the procedural fairness of the model, for a procedural-unfair model, we proposed a method to identify the features that lead to the procedural unfairness (i.e., unfair features) and proposed two methods to improve the procedural fairness of the model based on the detected unfair features. Experiments on nine datasets indicate that our method can accurately detect the features that lead to the procedural unfairness of the model, and both proposed mitigation methods can significantly improve the procedural fairness of the model while also increasing the distributive fairness, with only a slight decrease in the model's performance. Overall, the work presented in this paper significantly advances the understanding and methodologies for enhancing the procedural fairness of ML models.

In future research, there are several aspects worthy of further investigation: 

\begin{enumerate}
    \item \textbf{Enhancing Counterfactual Matching in Sparse Datasets:} As discussed in Section \ref{sec:limitation}, \textit{GPF\textsubscript{FAE}} metric requires sampling $n$ pairs of similar data points with different sensitive attribute values. In sparse datasets, such matching can be unreliable. We proposed an initial solution using KDE-based counterfactual generation to address this limitation. Future research could explore more advanced or data-efficient sample generation techniques to further improve the reliability of fairness assessment, or develop alternative methodologies that do not depend on the availability of similar data points at all.
    
    \item \textbf{Integrating Procedural Fairness into Training:} It would be worthwhile to explore the utilization of procedural fairness metrics to guide the training of ML models, so as to improve the procedural fairness of the models during the training process.
    
    \item \textbf{Extending to Individual Procedural Fairness:} Although this paper gives the definition of individual and group procedural fairness of ML models, it only studies the group procedural fairness in depth. Future work should explore how to quantify and improve individual procedural fairness in ML models.

    \item \textbf{Validating Generalizability and Real-World Applicability:} Since \textit{GPF\textsubscript{FAE}} is a newly proposed metric, its generalizability and practical value across diverse scenarios still require further empirical validation. In future work, we plan to systematically evaluate the adaptability and robustness of \textit{GPF\textsubscript{FAE}} across various sensitive attributes and multi-group fairness settings, as well as to explore its potential as a reference metric aligned with emerging AI regulatory and standardization frameworks.

    \item \textbf{Adapting \textit{GPF\textsubscript{FAE}} for Sequential and Time-Series Tasks:} Applying \textit{GPF\textsubscript{FAE}} (or its variants) to time series or sequential decision-making tasks is a promising area for future expansion. Such scenarios involve time dependency and require adaptive adjustments to attribution techniques and fairness definitions.

    \item \textbf{Balancing Fairness and Model Performance:} Prior research has underscored the existence of conflicts between the model performance and the distributive fairness~\cite{caton2020fairness,friedler2019comparative,speicher2018unified}. In this paper, we reveal that there is also a conflict between procedural fairness and model performance. Therefore, how to trade-off and comprehensively consider the model performance, procedural fairness, and distributive fairness is a challenging and desirable task for future research.
\end{enumerate}

\begin{acks}
This work was supported by the National Natural Science Foundation of China (Grant No. 62250710682), the Guangdong Provincial Key Laboratory (Grant No. 2020B121201001), the Program for Guangdong Introducing Innovative and Entrepreneurial Teams (Grant No. 2017ZT07X386), and an internal grant of Lingnan University.
\end{acks}

\printbibliography

\end{document}